\documentclass[10pt,twocolumn,letterpaper]{article}

\usepackage{cvpr}
\usepackage{times}
\usepackage{epsfig}
\usepackage{graphicx}
\usepackage{amsmath}
\usepackage{amssymb}

\usepackage{tabularx}
\usepackage{multirow}
\usepackage{multicol}

\usepackage{subfigure}

\usepackage[english]{babel}
\usepackage[T1]{fontenc}
\usepackage[utf8]{inputenc}

\usepackage{slashbox}

\usepackage{booktabs} % To thicken table lines

\usepackage{floatrow}
\newfloatcommand{capbtabbox}{table}[][\FBwidth]

\usepackage[pagebackref=true,breaklinks=true,letterpaper=true,colorlinks,bookmarks=false]{hyperref}

\cvprfinalcopy % *** Uncomment this line for the final submission

\def\cvprPaperID{1032} % *** Enter the CVPR Paper ID here

\ifcvprfinal\fi
\begin{document}

%%%%%%%%% TITLE

\title{Where am I looking at? Joint Location and Orientation Estimation by Cross-View Matching}

% \author{%
% Yujiao Shi, Xin Yu, Dylan Campbell and Hongdong Li\\
% Australian National University\\
% {\tt\small firstname.lastname@anu.edu.au}
% }

\author{Yujiao Shi,\textsuperscript{\rm 1, 2}~~
Xin Yu,\textsuperscript{\rm 1, 2, 3}~~
Dylan Campbell,\textsuperscript{\rm 1, 2}~~
Hongdong Li\textsuperscript{\rm 1, 2}
\\
\textsuperscript{\rm 1}Australian National University~~
\textsuperscript{\rm 2}Australian Centre for Robotic Vision~~
\textsuperscript{\rm 3}University of Technology Sydney\\
{\tt\small firstname.lastname@anu.edu.au}
% \textsuperscript{\rm 1}Australian National University, Canberra, Australia.\\
% \textsuperscript{\rm 2}Australian Centre for Robotic Vision, Australia.\\
% \textsuperscript{\rm 3}University of Technology Sydney.\\
% \{firstname.lastname\}@anu.edu.au
}

\maketitle
% \thispagestyle{empty}

%%%%%%%%% ABSTRACT
\begin{abstract}

Cross-view geo-localization is the problem of estimating the position and orientation (latitude, longitude and azimuth angle) of a camera at ground level given a large-scale database of geo-tagged aerial (\eg, satellite) images.
Existing approaches treat the task as a pure location estimation problem by learning discriminative feature descriptors, but neglect orientation alignment. 
It is well-recognized that knowing the orientation between ground and aerial images can significantly reduce matching ambiguity between these two views, especially when the ground-level images have a limited Field of View (FoV) instead of a full field-of-view panorama.
Therefore, we design a Dynamic Similarity Matching network to estimate cross-view orientation alignment during localization. 
In particular, we address the cross-view domain gap by applying a polar transform to the aerial images to approximately align the images up to an unknown azimuth angle.
Then, a two-stream convolutional network is used to learn deep features from the ground and polar-transformed aerial images.
Finally, we obtain the orientation by computing the correlation between cross-view features, which also provides a more accurate measure of feature similarity, improving location recall.
Experiments on standard datasets demonstrate that our method significantly improves state-of-the-art performance.
Remarkably, we improve the top-1 location recall rate on the CVUSA dataset by a factor of  $1.5\times$ for panoramas with known orientation, 
by a factor of  $3.3\times$ for panoramas with unknown orientation, and by a factor of  $6\times$ for $180^{\circ}$-FoV images with unknown orientation.
\end{abstract}

%%%%%%%%% BODY TEXT
\section{Introduction}

\begin{figure}[t]
    \centering
    \includegraphics[width=0.9\linewidth]{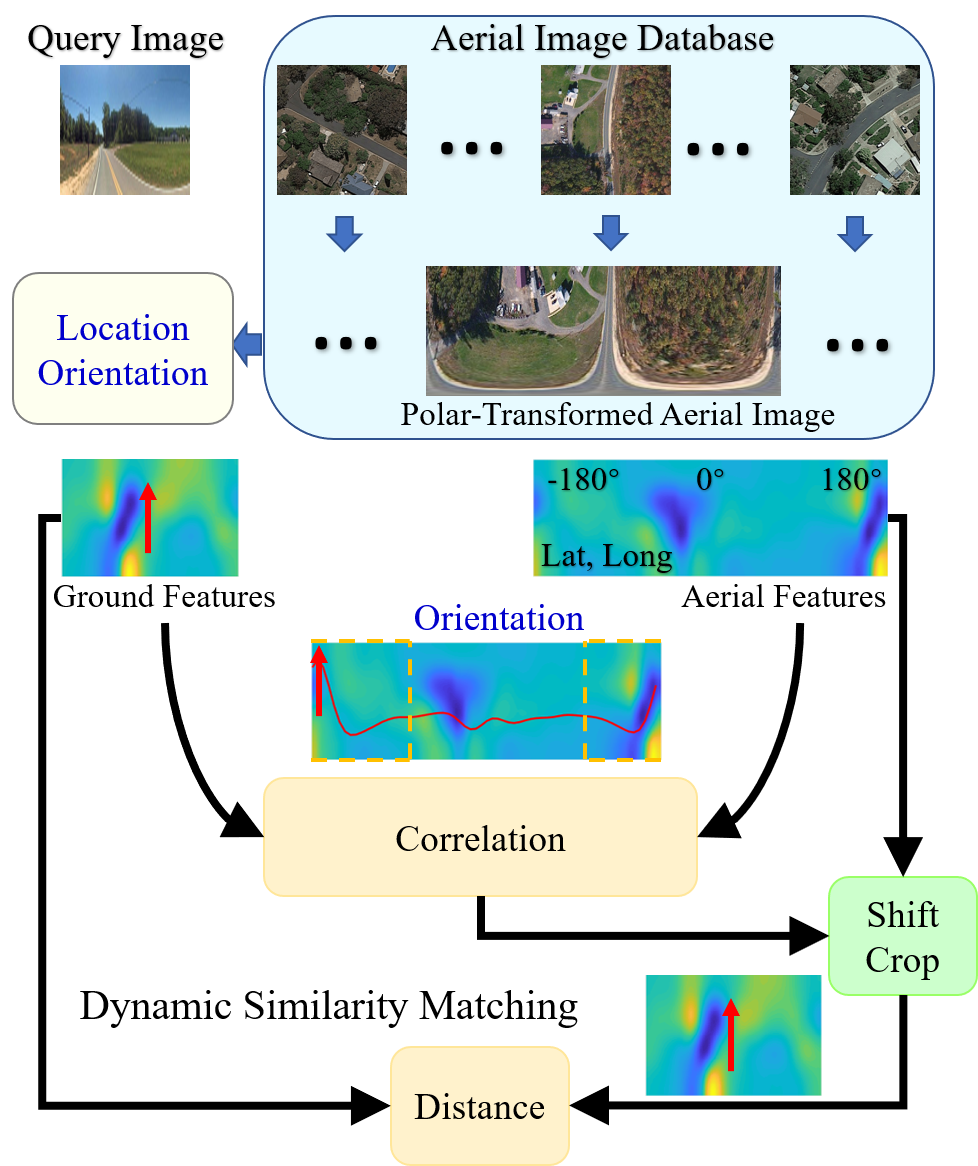}
    \vspace{-3mm}
    \caption{ \small
    Given a query ground image, we aim to recover its geographical location (latitude and longitude) and orientation (azimuth angle) by matching it against a large database of geo-tagged aerial images.
    To do so, we first apply a polar transform to aerial images to approximately align them with the ground view images, up to an unknown azimuth angle.
    We next compute the correlation between ground and aerial features along the axis of azimuth angles (horizontal direction) to estimate the orientation alignment of the ground image with respect to the aerial image.
    We then shift and crop out the corresponding aerial features, and compute their distance to the ground features, which provides a similarity measure for location retrieval.
    }
    \label{fig: openfigure}
\end{figure}
Given an image captured by a camera at ground level, it is reasonable to ask: where is the camera and which direction is it facing?
Cross-view image geo-localization aims to determine the geographical location and azimuth angle of a query image by matching it against a large geo-tagged satellite map covering the region.
Due to the accessibility and extensive coverage of satellite imagery, ground-to-aerial image alignment is becoming an attractive proposition for solving the image-based geo-localization problem.

However, cross-view alignment remains very difficult due to the extreme viewpoint change between ground and aerial images. The challenges are summarized as follows.
\vspace{-0.5em}
\begin{itemize}
\item[(1)] Significant visual differences between the views, including the appearance and projected location of objects in the scene, result in a large domain gap.
\vspace{-0.5em}
\item[(2)] The unknown relative orientation between the images, when the northward direction is not known in both, leads to localization ambiguities and increases the search space.
\vspace{-0.5em}
\item[(3)] Standard cameras have a limited Field of View (FoV), which reduces the discriminativeness of the ground-view features for cross-view localization, since the image region only covers local information and may match multiple aerial database images.
\vspace{-0.5em}
\end{itemize}

Existing methods cast this task as a pure location estimation problem and use deep metric learning techniques to learn viewpoint invariant features for matching ground and aerial images. 
Many approaches require the orientation to be provided, avoiding the ambiguities caused by orientation misalignments \cite{shi2019optimal, Regmi_2019_ICCV, Liu_2019_CVPR}.
However, the orientation is not always available for ground images in practice.
To handle this, some methods directly learn orientation invariant features \cite{vo2016localizing, Hu_2018_CVPR, Cai_2019_ICCV}, however they fail to address the large domain gap between ground and aerial images, limiting their localization performance.

\begin{figure}[t!]
    \centering
    \subfigure[Aerial]{
    \begin{minipage}{0.15\linewidth}
    \includegraphics[width=\linewidth, height=\linewidth]{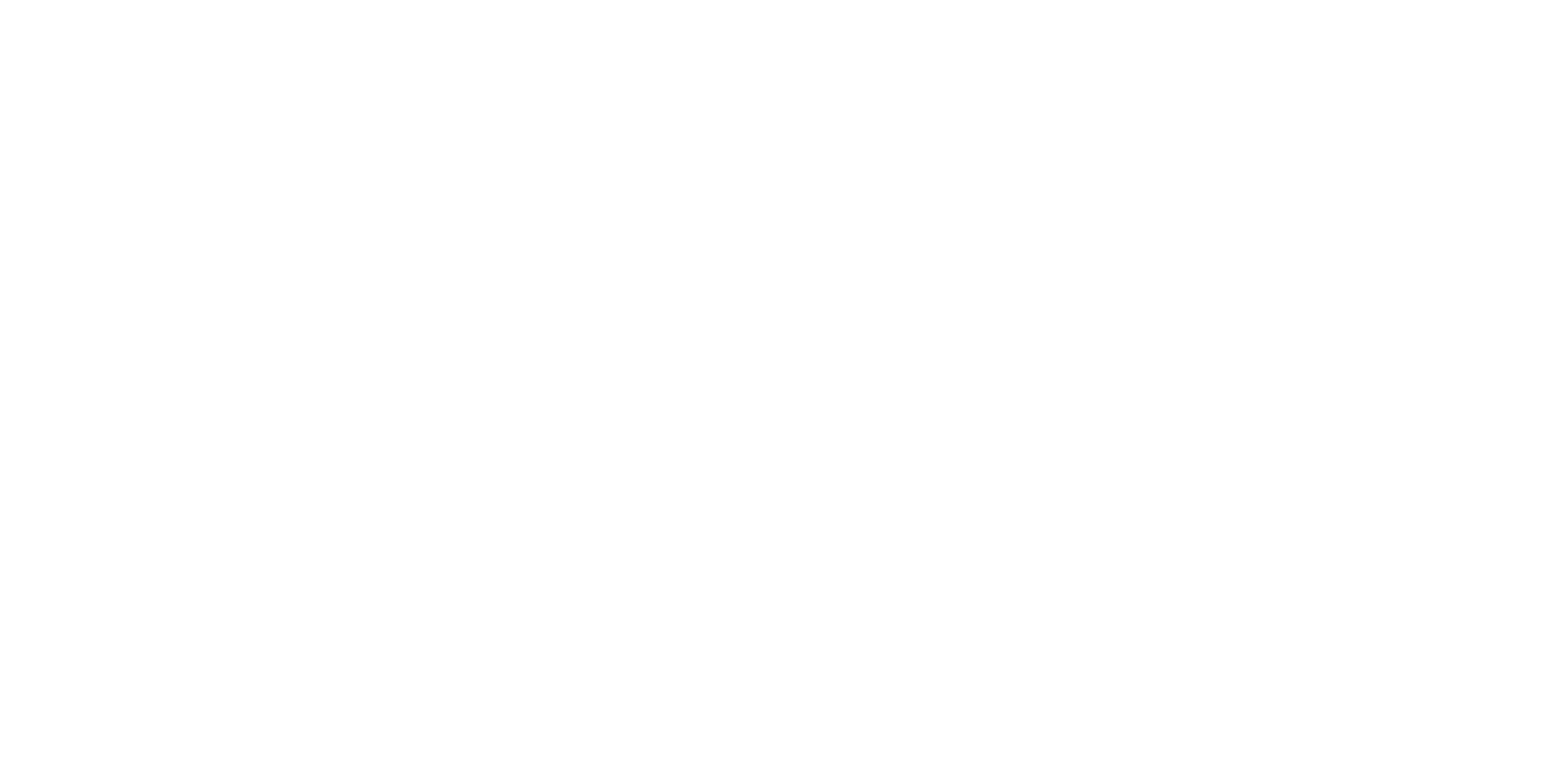}
    \includegraphics[width=\linewidth]{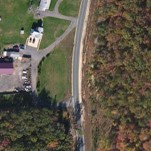}
    \end{minipage}
    }
    \hspace{-2mm}
    \subfigure[Ground]{
    \begin{minipage}{0.75\linewidth}
    \includegraphics[width=\linewidth, height=0.2\linewidth]{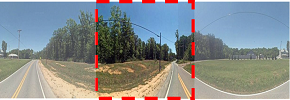}
    \includegraphics[width=\linewidth, height=0.2\linewidth]{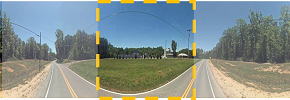}
    \end{minipage}
    \label{fig: challenge_orien_grd}
    }\\
    \vspace{-1em}
    \begin{minipage}{0.16\linewidth}
    \includegraphics[width=\linewidth, height=0.6\linewidth]{white_img.jpg}
    \end{minipage}
    \hspace{-2mm}
    \setcounter{subfigure}{2}
    \subfigure[Polar-transformed Aerial]{
    \begin{minipage}{0.75\linewidth}
    \includegraphics[width=\linewidth, height=0.2\linewidth]{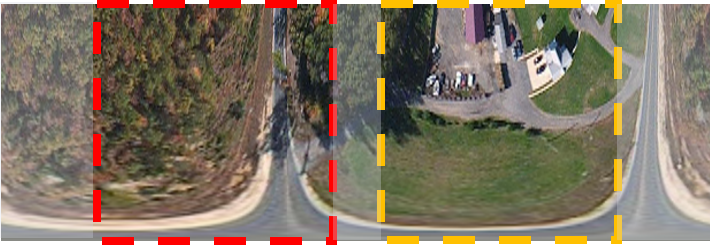}
    \end{minipage}
    \label{fig: challenge_FoV_grd}
    }\\
    \vspace{-3mm}
    \caption{\small The challenge of cross-view image matching caused by the unknown orientation and limited FoV of the query ground image. The scene content in panoramas captured at the same location but with different azimuth angles (top and middle) is offset, and the image content in a limited FoV image can be entirely different to another image captured from the same location.}
    \label{fig:illustration of unknwon orien and limited FoV}
\end{figure}

To reduce the cross-view domain gap, we explore the geometric correspondence between ground and aerial images. We observe that there are two geometric cues that are statistically significant in real ground images under an equirectangular projection:
(i) horizontal lines in the image (parallel to the azimuth axis) have approximately constant depth and so correspond to concentric circles in the aerial image; and
(ii) vertical lines in the image have depth that increases with the $y$ coordinate and so correspond to radial lines in the aerial image.
To be more specific, if the scene was flat, then a horizontal line in the ground image maps to a circle in the aerial image.
We make use of these geometric cues by applying a polar coordinate transform to the aerial images, mapping concentric circles to horizontal lines. This reduces the differences in the projected geometry and hence the domain gap, as shown in Figure~\ref{fig:illustration of unknwon orien and limited FoV}.

We then employ a two-stream CNN to learn feature correspondences between ground and aerial images. We extract feature volumes that preserve the spatial relationships between features, which is a critical cue for geo-localization.
However, orientation misalignments lead to inferior results when using spatially-aware image features.
Moreover, it is difficult to match features when a limited FoV is imaged, since the ground image contains only a small sector of the aerial image.
Therefore, our intuition is to find the orientation alignment and thus facilitate accurate similarity matching.

As the polar transform projects the aerial image into the ground-view camera coordinate frame, it allows to estimate the orientation of each ground image with respect to its aerial counterpart by feature correlation. 
In this paper, we propose a Dynamic Similarity Matching (DSM) module to achieve that goal.
% estimate the orientation of each ground image with respect to its aerial counterpart. 
% Thanks to the polar transform, the horizontal lines in the transformed aerial images corresponds the azimuth direction. This allows to estimate orientation of ground images with respect to aerial images by using a Dynamic Similarity Matching (DSM) module.
To be specific, we compute the correlation between the ground and aerial features in order to generate a similarity score at each angle, marked by the red curve in Figure \ref{fig: openfigure}. The position of the similarity score maximum corresponds to the latent orientation of the ground image with respect to the aerial image. If the ground image has a limited FoV, we extract the appropriate local region from the aerial feature representation for localization.
By using our DSM module, the feature similarity between the ground and aerial images is measured more accurately. Therefore, our method outperforms the state-of-the-art by a large margin. 
% Notably, we improve the top-1 location recall rate on the CVUSA dataset \cite{zhai2017predicting} by a factor of $1.5\times$ for panoramas with known orientation, by a factor of $3.3\times$ for panoramas with unknown orientation, and by a factor of $6\times$ for $180^{\circ}$-FoV images with unknown orientation.

The contributions of our work are:
\vspace{-0.5em}
\begin{itemize}
\item the first image-based geo-localization method to jointly estimate the position and orientation\footnote{Throughout this paper, orientation refers to the 1-DoF azimuth angle.} of a query ground image regardless of its Field of View;
\vspace{-0.5em}
\item a Dynamic Similarity Matching (DSM) module to measure the feature similarity of the image pair while accounting for the orientation of the ground image, facilitating accurate localization; and
\vspace{-0.5em}
\item extensive experimental results demonstrating that our method achieves significant performance improvements over the state-of-the-art in various geo-localization scenarios.
\end{itemize}

%-------------------------------------------------------------------------
\section{Related Work}

\begin{figure*}[h]
    \centering
    \includegraphics[width=0.85\linewidth]{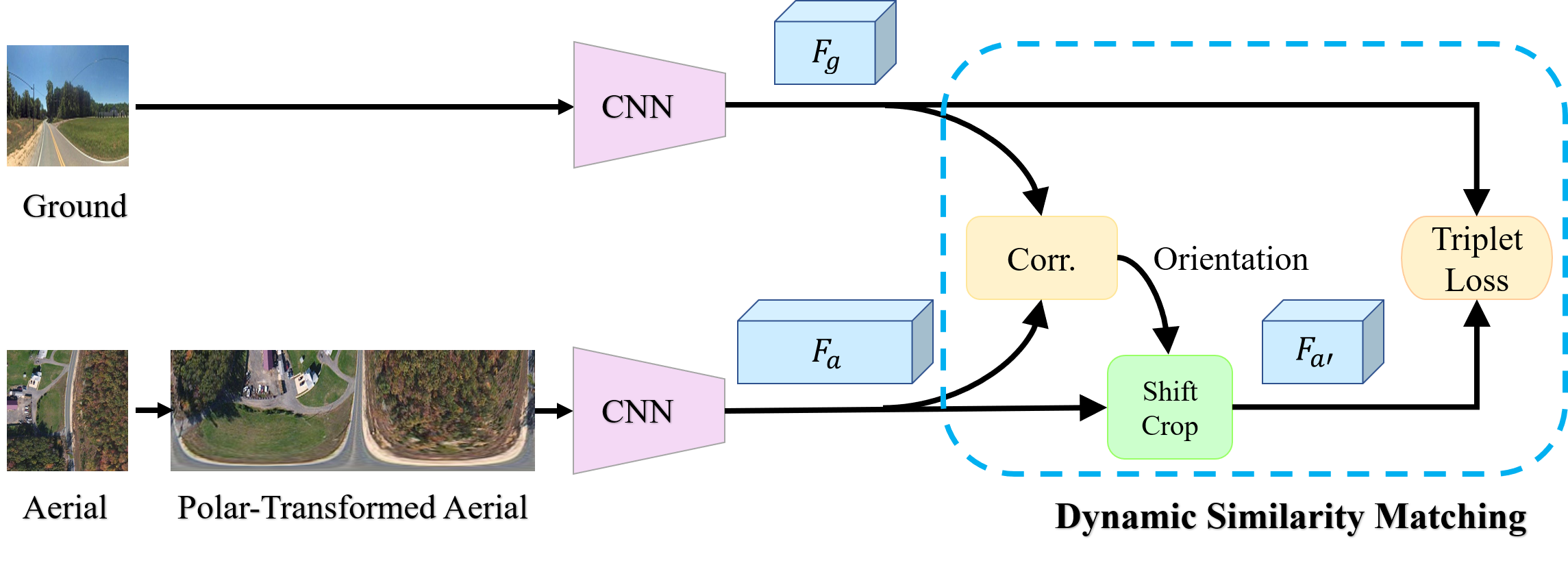}
    \vspace{-2mm}
    \caption{\small
    Flowchart of the proposed method. A polar transform is first applied to the aerial image, and then a two-stream CNN is employed to extract features from ground and polar-transformed aerial images. Given the extracted feature volume representations, the correlation between the two is used to estimate the orientation of the ground image with respect to the aerial image. Next, the aerial features are shifted and cropped to obtain the section that (potentially) corresponds to the ground features. The similarity of the resulting features is then used for location retrieval.
    }
    \label{fig:framework}
    \vspace{-2mm}
\end{figure*}

Existing cross-view image-based geo-localization aims to estimate the location (latitude and longitude) of a ground image by matching it against a large database of aerial images. Due to the significant viewpoint changes between ground and aerial images, hand-crafted feature matching \cite{castaldo2015semantic, lin2013cross, mousavian2016semantic} becomes the bottleneck of the performance of cross-view geo-localization. Deep convolutional neural networks (CNNs) have proven their powerful capability on image representations \cite{russakovsky2015imagenet}. This motivates recent geo-localization works to extract features from ground and aerial images with CNNs.

Workman and Jacobs \cite{workman2015location} first introduced deep features to the cross-view matching task. They used an AlexNet \cite{krizhevsky2012imagenet} network fine-tuned on Imagenet \cite{russakovsky2015imagenet} and Places \cite{zhou2014learning} to extract deep features for cross-view image matching. They demonstrated that further tuning of the aerial branch by minimizing the distance between matching ground and aerial pairs led to better localization performance \cite{workman2015wide}.
% \textbf{Learning powerful descriptors:}
Vo and Hays \cite{vo2016localizing} investigated a set of CNN architectures (classification, hybrid, Siamese and triplet CNNs) for matching cross-view images. Considering the orientation misalignments between ground and aerial images, they proposed an auxiliary orientation regression block to let the network learn orientation-aware feature representations, and used multiple aerial images with different orientations during testing phase. To learn orientation invariant features, Hu \etal\cite{Hu_2018_CVPR} embedded a NetVlad layer \cite{arandjelovic2016netvlad} on top of a two-branch CNN for cross-view image matching. Cai \etal \cite{Cai_2019_ICCV} introduced a lightweight attention module to reweight spatial and channel features to obtain more representative descriptors, and then proposed a hard exemplar reweighting triplet loss to improve the quality of network training. They also employed an orientation regression block to force the network to learn orientation-aware features. Sun \etal \cite{sun2019geocapsnet} employed capsule networks to encode spatial feature hierarchies for feature representations. Although these methods learned orientation-aware descriptors for localization, they overlooked the domain difference between ground and aerial images. 

To bridge the large domain gap between ground and aerial images, 
Zhai \etal \cite{zhai2017predicting} learned a transformation matrix between aerial and ground features for predicting ground semantic information from aerial images.
Regmi and Shah \cite{Regmi_2019_ICCV} synthesized an aerial image from a ground one using a generative model, and then fused features of the ground image and the synthesized aerial image as the descriptor for retrieval.
Shi \etal \cite{shi2019optimal} proposed a feature transport module to map ground features into the aerial domain and then conducted similarity matching. 
Shi \etal \cite{shi2019spatial} also used polar transform to first bridge the geometric domain difference and then a spatial-aware feature aggregation module to select salient features for global feature descriptor representation. 
% As these methods involve generative procedures from one domain to another, they require the ground images to be panoramas or orientation-aligned. 
However, all those methods require ground images to be panoramas or orientation-aligned.
Finally, Liu \& Li \cite{Liu_2019_CVPR} found that the orientation provided important clues for determining the location of a ground image, and thus explicitly encoded the ground-truth orientation as an additional network input.

In contrast to existing works, we aim to estimate the location and orientation of ground images jointly, since exploring orientation information can facilitate cross-view matching for both panoramas and images with limited FoV.

\section{Location and Orientation Estimation by Cross-view Image Matching}

In the cross-view image-based geo-localization task, ground images are captured by a camera whose image plane is perpendicular to the ground plane and $y$ axis is parallel to the gravity direction, and aerial images are captured from a camera whose image plane is parallel to the ground plane. 
Since there are large appearance variations between these two image domains, our strategy is to first reduce the projection differences between the viewpoints and then to extract discriminative features from the two domains.
Furthermore, inspired by how humans localize themselves, we use the spatial relationships between objects as a critical cue for inferring location and orientation. Therefore, we enable our descriptors to encode the spatial relationship among the features, as indicated by $F_g$ and $F_a$ in Figure~\ref{fig:framework}.

Despite the discriminativeness of the spatially-aware features, they are very sensitive to orientation changes. For instance, when the azimuth angle of a ground camera changes, the scene contents will be offset in the ground panorama, and the image content may be entirely different if the camera has a limited FoV, as illustrated in Figure~\ref{fig:illustration of unknwon orien and limited FoV}. Therefore, finding the orientation of the ground images is crucial to make the spatially-aware features usable. To this end, we propose a dynamic similarity matching (DSM) module, as illustrated in Figure~\ref{fig:framework}. With this module, we not only estimate the orientation of the ground images but also achieve more accurate feature similarity scores, regardless of orientation misalignments and limited FoVs, thus enhancing geo-localization performance.

\subsection{A Polar Transform to Bridge the Domain Gap}
Since ground panoramas\footnote{Although we use panoramic images as an example, the correspondence relationships between ground and aerial images also apply to images with limited FoV.}
project 360-degree rays onto an image plane using an equirectangular projection, and are orthogonal to the satellite-view images, vertical lines in the ground image correspond to radial lines in the aerial image, and horizontal lines correspond approximately to circles in the aerial image, assuming that the pixels along the line have similar depths, which occurs frequently in practice.
This layout correspondence motivates us to apply a polar transform to the aerial images.
In this way, the spatial layouts of these two domains can be roughly aligned, as illustrated in Figure \ref{fig: challenge_orien_grd} and Figure \ref{fig: challenge_FoV_grd}.

To be specific, the polar origin is set to the center of each aerial image, corresponding to the geo-tag location, and the $0^{\circ}$ angle is chosen as the northward direction, corresponding to the upwards direction of an aligned aerial image. 
In addition, we constrain the height of the polar-transformed aerial images to be the same as the ground images, and ensure that the angle subtended by each column of the polar transformed aerial images is the same as in the ground images.
We apply a uniform sampling strategy along radial lines in the aerial image, 
such that the innermost and outermost circles of the aerial image are mapped to the bottom and top line of the transformed image respectively.

Formally, let $S_a \times S_a$ represent the size of an aerial image and $H_g \times W_g$ denote the target size of polar transform. The polar transform between the original aerial image points $(x_i^a, y_i^a)$ and the target polar transformed ones $(x_i^t, y_i^t)$ is
\begin{equation}
\small
 \label{PT}
\begin{split}
x_{i}^{a} & = \frac{S_{a}}{2} - \frac{S_{a}}{2}\frac{(H_g - x_i^t)}{H_g} \cos \left(\frac{2 \pi}{W_{g}}y_i^t\right),\\
y_{i}^{a} & = \frac{S_{a}}{2} + \frac{S_{a}}{2}\frac{(H_g - x_i^t)}{H_g} \sin \left(\frac{2 \pi}{W_{g}}y_i^t\right).\\
\end{split}
\end{equation}
By applying a polar transform, we coarsely bridge the projective geometry domain gap between ground and aerial images. 
This allows the CNNs to focus on learning the feature correspondences between the ground and polar-transformed aerial images without consuming network capacity on learning the geometric relationship between these two domains.

\subsection{A Spatially-Aware Feature Representation}
Applying a translation offset along the $x$ axis of a polar-transformed image is equivalent to rotating the aerial image.
Hence, the task of learning rotational equivariant features for aerial images becomes learning translational equivariant features, which significantly reduces the learning difficulty for our network since CNNs inherently have the property of translational equivariance \cite{lenc2015understanding}.
However, since the horizontal direction represents a rotation, we have to ensure that the CNN treats the leftmost and rightmost columns of the transformed image as adjacent.
Hence, we propose to use circular convolutions with wrap-around padding along the horizontal direction.

We adopt VGG16 \cite{Simonyan2014VeryDC} as our backbone network.
In particular, the first ten layers of VGG16 are used to extract features from the ground and polar-transformed aerial images.
Since the polar transform might introduce distortions along the vertical direction,
due to the assumption that horizontal lines have similar finite depths,
we modify the subsequent three layers which decrease the height of the feature maps but maintain their width.
In this manner, our extracted features are more tolerant to distortions along the vertical direction while retaining information along the horizontal direction.
We also decrease the feature channel number to $16$ by using these three convolutional layers, and obtain a feature volume of size $4\times64\times16$.
Our feature volume representation is a global descriptor designed to preserve the spatial layout information of the scene, thus increasing the discriminativeness of the descriptors for image matching.

\subsection{Dynamic Similarity Matching (DSM)} 
When the orientation of ground and polar-transformed aerial features are aligned, their features can be compared directly.
However, the orientation of the ground images is not always available, and orientation misalignments increase the difficulty of geo-localization significantly, especially when the ground image has a limited FoV. 
When humans are using a map to relocalize themselves, they determine their location and orientation jointly by comparing what they have seen with what they expect to see on the map. 
% In contrast, humans are not affected by orientation misalignments when they use a map to relocalize themselves.
% They determine their location and orientation jointly by comparing what they have seen with what they expect to see on the map.
In order to let the network mimic this process, we compute the correlation between the ground and aerial features along the azimuth angle axis.
Specifically, we use the ground feature as a sliding window and compute the inner product between the ground and aerial features across all possible orientations. 
Let $F_a \in R^{H \times W_a \times C}$ and $F_g \in R^{H \times W_g \times C}$ denote the aerial and ground features respectively, where $H$ and $C$ indicate the height and channel number of the features, $W_a$ and $W_g$ represent the width of the aerial and ground features respectively, and $W_a \geq W_g$. The correlation between $F_a$ and $F_g$ is expressed as
% \vspace{-0.5em}
\begin{equation}
\small
    [F_a * F_g](i) \! = \!\!\sum_{c=1}^{C} \sum_{h=1}^{H} \sum_{w=1}^{W_{g}} \! F_a(h, (i+w) \% W_a, c) F_g(h, w, c),
 \label{Eq: corr}
\end{equation}
where $F(h, w, c)$ is the feature response at index ($h$, $w$, $c$), and $\%$ denotes the modulo operation.
After correlation computation, the position of the maximum value in the similarity scores is the estimated orientation of the ground image with respect to the polar-transformed aerial one.

When a ground image is a panorama, regardless of whether the orientation is known, the maximum value in the correlation results is directly converted to the $L_2$ distance by computing $2(1-\max ([F_a * F_g](i))$, where $F_a$ and $F_g$ are $L_2$-normalized.  
When a ground image has a limited FoV, we crop the aerial features corresponding to the FoV of the ground image at the position of the maximum similarity score.
Then we re-normalize the cropped aerial features and calculate the $L_2$ distance between the ground and aerial features as the similarity score for matching.
Note that if there are multiple maximum similarity scores, we choose one randomly, since this means that the aerial images has symmetries that cannot be disambiguated.

\subsection{Training DSM} 

During the training process, our DSM module is applied to all ground and aerial pairs, whether they are matching or not. 
% For matching pairs, DSM forces the network to learn feature embeddings of ground and polar-transformed aerial images to be similar and to be discriminative along the horizontal direction (\ie, azimuth).
For matching pairs, DSM forces the network to learn similar feature embeddings for ground and polar-transformed aerial images with discriminative feature representations along the horizontal direction (\ie, azimuth).
In this way, DSM is able to identify the orientation misalignment as well as find the best feature similarity for matching.
For non-matching pairs, as it is the most challenging case when they are aligned (\ie, their similarity is larger), our DSM is also used to find the most feasible orientation for a ground image aligning to a non-matching aerial one,
% For non-matching pairs, DSM is also used to find the most feasible orientation for a ground image aligning to a non-matching aerial one.
and we minimize the maximum similarity of non-matching pairs to make the features more discriminative. 
Following traditional cross-view localization methods \cite{Hu_2018_CVPR, Liu_2019_CVPR, shi2019optimal}, we employ the weighted soft-margin triplet loss \cite{Hu_2018_CVPR} to train our network
\begin{equation}
\small
     \mathcal{L} = \log \left ( 1 + e^{\alpha\left ( \left \| F_g - F_{a^{'}} \right \|_F - \left \| F_g - F_{a^{\ast'}} \right \|_F \right ) } \right ),
     \label{eq: triplet_loss}
\end{equation}
where $F_g$ is the query ground feature, $F_{a^{'}}$ and $F_{a^{\ast'}}$ indicate the cropped aerial features from the matching aerial image and a non-matching aerial image respectively, and $\left \|\cdot \right \|_F$ denotes the Frobenius norm. The parameter $\alpha$ controls the convergence speed of training process; following precedents we set it to $10$ \cite{Hu_2018_CVPR, Liu_2019_CVPR, shi2019optimal}.

\subsection{Implementation Details}
We use the first ten convolutional layers in VGG16 with pretrained weights on Imagenet \cite{deng2009imagenet}, and randomly initialize the parameters in the following three layers for global feature descriptor extraction. The first seven layers are kept fixed and the subsequent six layers are learned. The Adam optimizer \cite{kingma2014adam} with a learning rate of $10^{-5}$ is employed for training. Following \cite{vo2016localizing, Hu_2018_CVPR, Liu_2019_CVPR, shi2019optimal}, we adopt an exhaustive mini-batch strategy \cite{vo2016localizing} with a batch size of $B=32$ to create the training triplets. Specifically, for each ground image within a mini-batch, there is one matching aerial image and $(B-1)$ non-matching ones. Thus we construct $B(B-1)$ triplets. Similarly, for each aerial image, there is one matching ground image and $(B-1)$ non-matching ones within a mini-batch, and thus we create another $B(B-1)$ triplets. Hence, we have $2B(B-1)$ triplets in total.

%-------------------------------------------------------------------------
\section{Experiments}
\subsection{Datasets} 
\begin{figure}
    \centering
    \includegraphics[width=0.17\linewidth, height=0.17\linewidth]{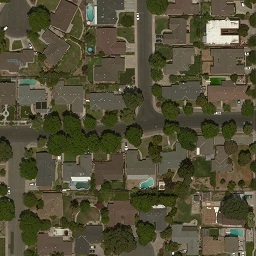}
    \includegraphics[width=0.75\linewidth, height=0.17\linewidth]{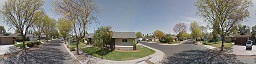}\\
    \includegraphics[width=0.17\linewidth, height=0.17\linewidth]{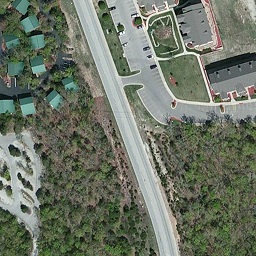}
    \includegraphics[width=0.75\linewidth, height=0.17\linewidth]{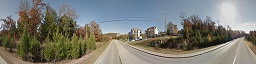}\\
    \includegraphics[width=0.17\linewidth, height=0.17\linewidth]{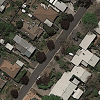}
    \includegraphics[width=0.75\linewidth, height=0.17\linewidth]{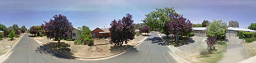}\\
    \includegraphics[width=0.17\linewidth, height=0.17\linewidth]{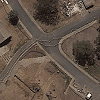}
    \includegraphics[width=0.75\linewidth, height=0.17\linewidth]{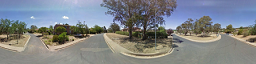}\\
    \vspace{-2mm}
    \caption{\small Cross-view image pairs from the CVUSA (top two rows) and CVACT (bottom two rows) datasets. The aerial images are on the left and the ground panoramas are on the right.}
    \label{fig:dataset}
\end{figure}

% frequently-used
We carry out the experiments on two standard cross-view datasets, CVUSA \cite{zhai2017predicting} and CVACT \cite{Liu_2019_CVPR}. They both contain $35,532$ training ground and aerial pairs and $8,884$ testing pairs. Following an established testing protocol \cite{Liu_2019_CVPR, shi2019optimal}, we denote the test sets in CVUSA and CVACT as CVUSA and CVACT\_val, respectively.
CVACT also provides a larger test set, CVACT\_test, which contains $92,802$ cross-view image pairs for fine-grained city-scale geo-localization.  Note that the ground images in both of the two datasets are panoramas, and all the ground and aerial images are north aligned. Figure \ref{fig:dataset} presents samples of cross-view image pairs from the two datasets.

Furthermore, we also conduct experiments on ground images with unknown orientation and limited FoV.
We use the image pairs in CVUSA and CVACT\_val, and randomly rotate the ground images along the azimuth direction and crop them according to a predetermined FoV. 
The constructed test set with different FoVs as well as our source code are available via https://github.com/shiyujiao/cross\_view\_localization\_DSM.git.

\subsection{Evaluation Metrics}

\noindent\textbf{Location estimation:}
Following the standard evaluation procedure for cross-view image localization \cite{vo2016localizing, Hu_2018_CVPR, Liu_2019_CVPR, shi2019optimal, Cai_2019_ICCV, sun2019geocapsnet, Regmi_2019_ICCV}, we use the top $K$ recall as the location evaluation metric to examine the performance of our method and compare it with the state-of-the-art.
Specifically, given a ground image, we retrieve the top $K$ aerial images in terms of $L_2$ distance between their global descriptors. The ground image is regarded as successfully localized if its corresponding aerial image is retrieved within the top $K$ list. The percentage of correctly localized ground images is recorded as recall at top $K$ (r@$K$).

% \medskip
\noindent\textbf{Orientation estimation:}
The predicted orientation of a query ground image is meaningful only when the ground image is localized correctly. 
Hence, we evaluate the orientation estimation accuracy of our DSM only on ground images that have been correctly localized by the top-1 recall. 
% That is, the orientation estimation accuracy is only measured for the top-1 correctly retrieved locations. 
In this experiment, when the differences between the predicted orientation of a ground image and its ground-truth orientation is within $\pm 10\%$ of its FoV, the orientation estimation of this ground image is deemed as a success. We record the percentage of ground images for which the orientation is correctly predicted as the orientation estimation accuracy (orien\_acc). 
Since aerial images are often rotationally symmetric, orientation estimation can yield large errors, such as 180$^{\circ}$ for scenes that look similar in opposite directions.
Hence we report the robust median orientation error, denoted as median\_error, instead of the mean.

\begin{table}[tbp]
% \footnotesize
% \setlength{\tabcolsep}{2pt}
\small
\centering
\vspace{-2mm}
\caption{\small Comparison with existing methods on the CVUSA \cite{zhai2017predicting} dataset. Here, ``--'' denotes that the results on the corresponding evaluation metric are not available as some of the works only use r@1\% as the evaluation metric. }
\begin{tabular}{c|c c c c}
\toprule
 \multirow{2}{*}{Methods} & \multicolumn{4}{c}{CVUSA}      \\ \cline{2-5} 
                                      & r@1   & r@5   & r@10  & r@1\%   \\\hline \hline
 Workman \etal \cite{workman2015wide}  & --     & --     & --     & 34.3 \\
 Zhai \etal \cite{zhai2017predicting}  & --     & --     & --     & 43.2   \\  
 Vo and Hays \cite{vo2016localizing}   & --     & --     & --    & 63.7   \\  
 CVM-NET \cite{Hu_2018_CVPR}        & 22.47 & 49.98 & 63.18 & 93.62  \\  
 Liu \& Li \cite{Liu_2019_CVPR}        & 40.79 & 66.82 & 76.36 & 96.12        \\ 
 Regmi and Shah \cite{Regmi_2019_ICCV}  & 48.75 & --     & 81.27 & 95.98        \\ 
 Siam-FCANet34 \cite{Cai_2019_ICCV}          & --     & --     & --     & 98.3        \\ 
 CVFT \cite{shi2019optimal}            & 61.43 & 84.69 & 90.49 & 99.02       \\ 
 Ours                  & \textbf{91.96} & \textbf{97.50} & \textbf{98.54} & \textbf{99.67}   \\ 
\bottomrule
\end{tabular}
\label{tab: compare_stoa_CVUSA}
\end{table}

\subsection{Localizing Orientation-Aligned Panoramas}

\begin{table}[tbp]
\small
% \footnotesize
% \setlength{\tabcolsep}{8pt}
\centering
\vspace{-2mm}
\caption{\small Comparison with existing methods on the CVACT\_val \cite{Liu_2019_CVPR} dataset by re-training existing networks using the code provided by the authors.}
\begin{tabular}{c | c c c c}
\toprule
 \multirow{2}{*}{Methods} & \multicolumn{4}{c}{CVACT\_val} \\ \cline{2-5} 
                                       & r@1   & r@5   & r@10  & r@1\% \\\hline \hline
 CVM-NET  \cite{Hu_2018_CVPR}          & 20.15 & 45.00 & 56.87 & 87.57    \\  
 Liu \& Li \cite{Liu_2019_CVPR}        & 46.96 & 68.28 & 75.48 & 92.01      \\ 
 CVFT \cite{shi2019optimal}            & 61.05 & 81.33 & 86.52 & 95.93      \\ 
 Ours                     & \textbf{82.49}  & \textbf{92.44}     & \textbf{93.99}     & \textbf{97.32}       \\ 
\bottomrule
\end{tabular}
\label{tab: compare_stoa_CVACT}
\end{table}

\begin{figure*}[t]
    \centering
    \subfigure[\small CVUSA]{
    \centering
    \scalebox{1.0}[1.0]{\includegraphics[trim={8mm 65mm 10mm 70mm}, clip, width=0.32\linewidth]{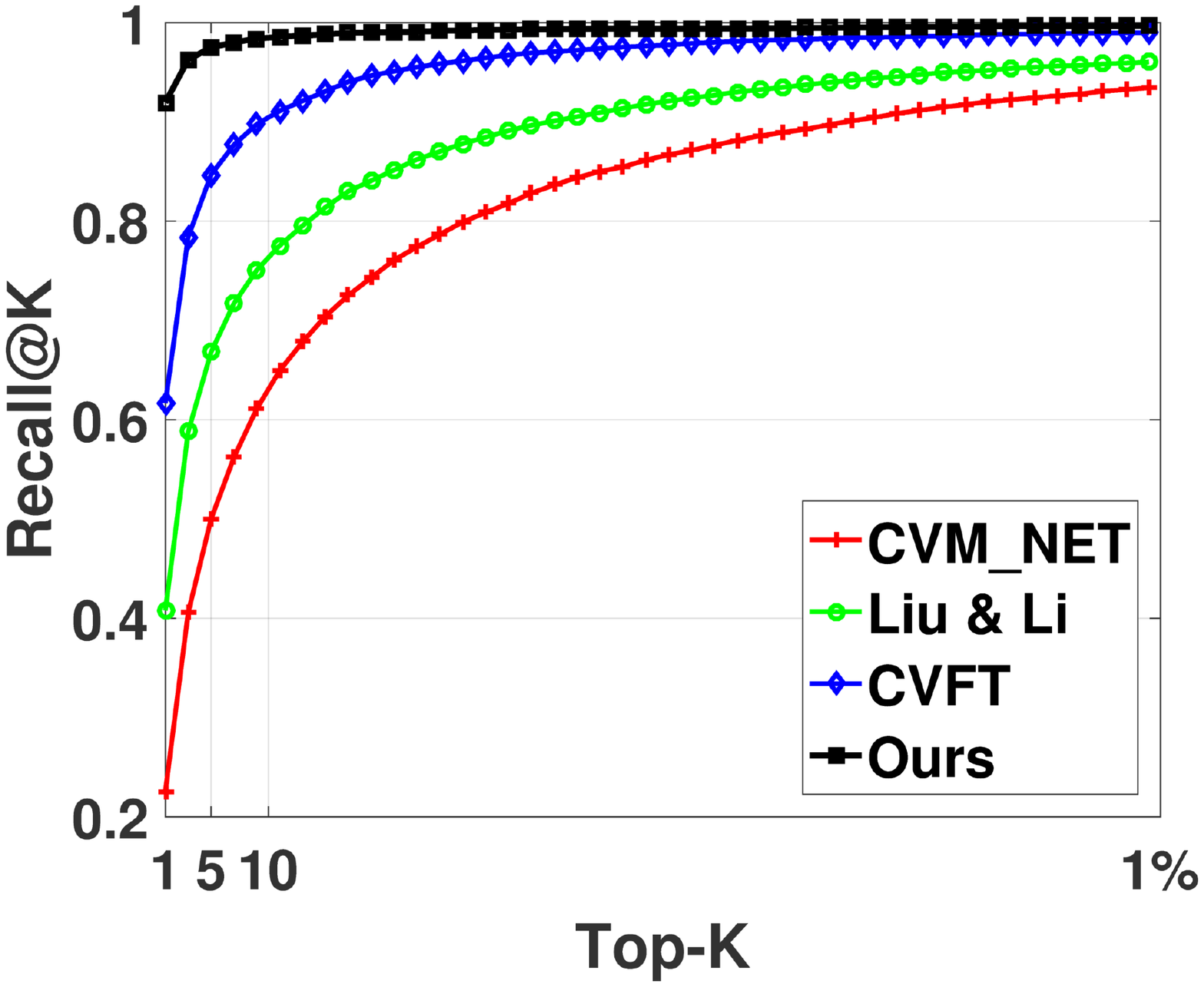}}
    \label{fig: recall_CVUSA}
    }
    \hspace{-6mm}
    \subfigure[\small CVACT\_val]{
    \centering
     \scalebox{1.0}[1.0]{\includegraphics[trim={8mm 65mm 10mm 70mm}, clip, width=0.32\linewidth]{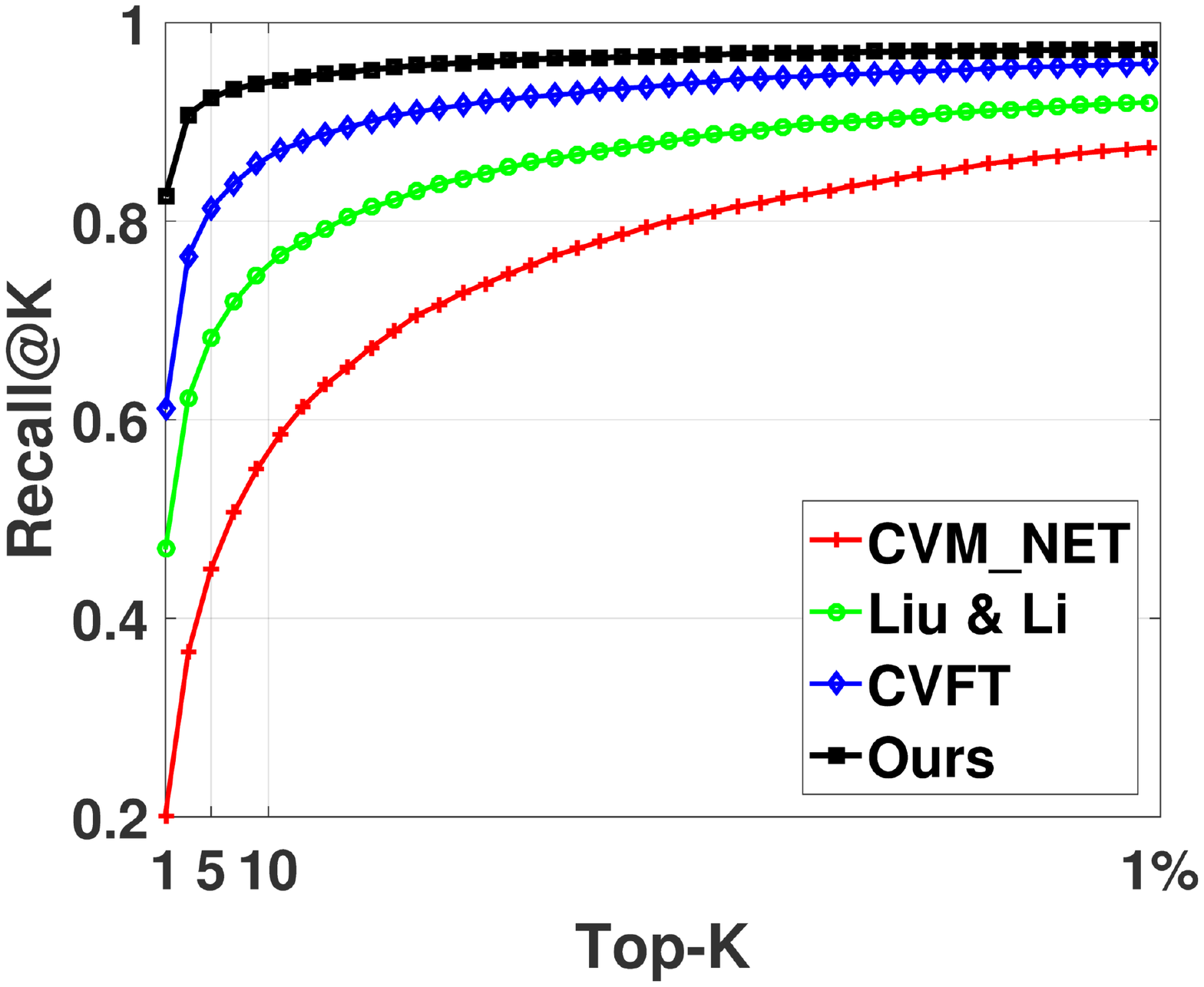}}
    \label{fig: recall_CVACT_val}
    }
     \hspace{-6mm}
    \subfigure[\small CVACT\_test]{
    \centering
    \scalebox{1.0}[1.0]{\includegraphics[trim={8mm 65mm 10mm 70mm}, clip, width=0.32\linewidth]{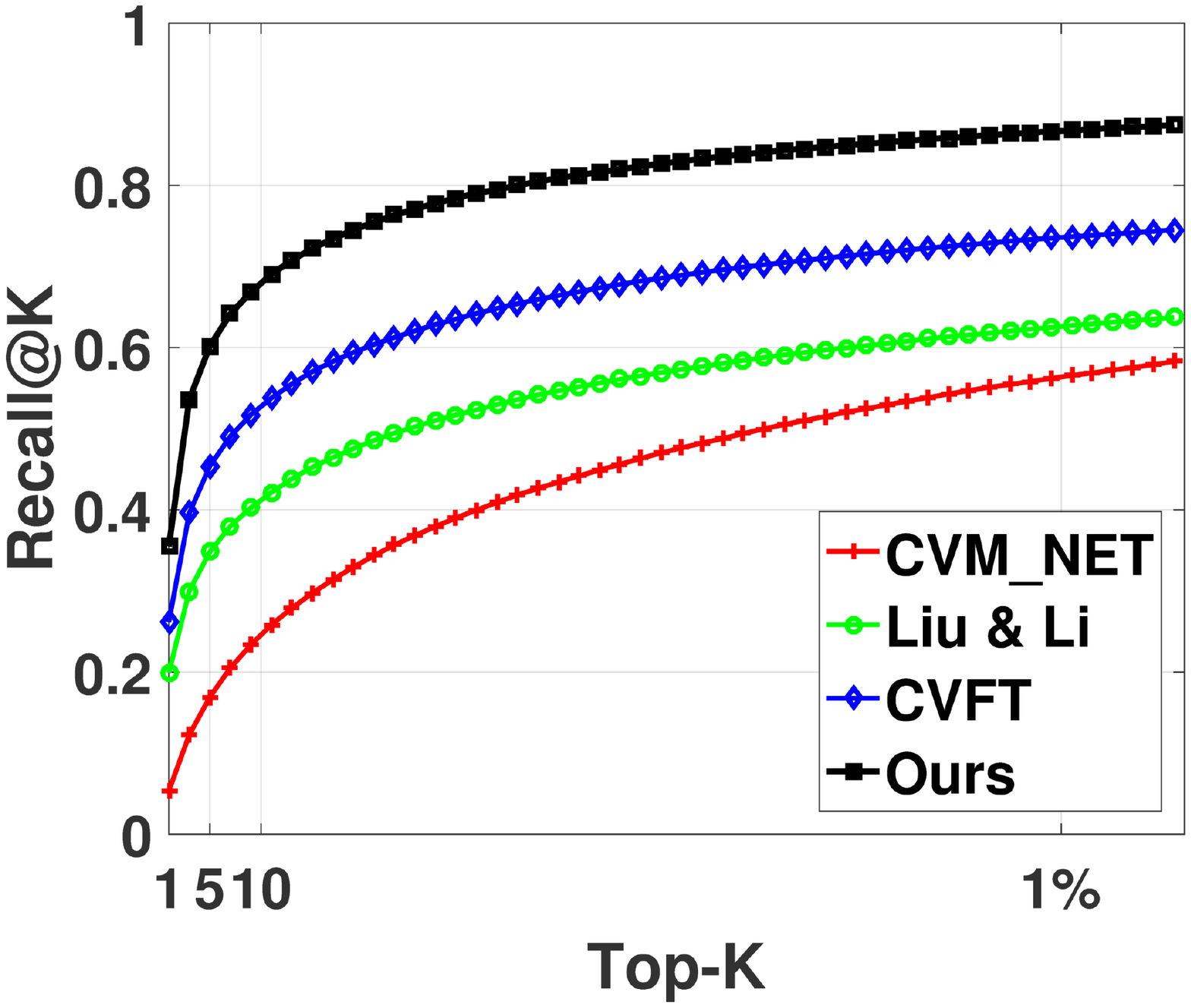}}
    \label{fig: recall_CVACT_test}
    }
    \vspace{-3mm}
    \caption{\small Recall comparison at different values of $K$ on the CVUSA, CVACT\_val and CVACT\_test datasets.}
    \vspace{-1mm}
    \label{fig:recall_CVUSA_CVACT}
\end{figure*}

We first investigate the location estimation performance of our method and compare it with the state-of-the-art on the standard CVUSA and CVACT datasets, where ground images are orientation-aligned panoramas. 
In Table \ref{tab: compare_stoa_CVUSA}, we present our results on the CVUSA dataset with the recall rates reported in other works \cite{Liu_2019_CVPR, shi2019optimal, Hu_2018_CVPR, Regmi_2019_ICCV, Cai_2019_ICCV}.
% As is clear in Table \ref{tab: compare_stoa_CVUSA}, our method significantly outperforms the state-of-the-art methods by a large margin, 
% achieving a $1.5\times$ improvement on r@1.
We also retrain existing networks \cite{Hu_2018_CVPR, Liu_2019_CVPR, shi2019optimal} on the CVACT dataset using source code provided by the authors.
The recall results at top-1, top-5, top-10 and top-1\% on CVACT\_val are presented in Table \ref{tab: compare_stoa_CVACT}, and the complete r@$K$ performance curves on CVUSA and CVACT\_val are illustrated in Figure \ref{fig: recall_CVUSA} and Figure \ref{fig: recall_CVACT_val}, respectively.

% It can be seen that our method achieves consistently better recall results than all the comparison algorithms, demonstrating the effectiveness of our method. 

Among those comparison methods, \cite{workman2015wide,zhai2017predicting, vo2016localizing} are first explorers to apply deep-based methods to cross-view related tasks.
CVM-NET \cite{Hu_2018_CVPR} and Siam-FCANet34 \cite{Cai_2019_ICCV} focus on designing powerful feature extraction networks. Liu \& Li \cite{Liu_2019_CVPR} introduce the orientation information to networks so as to facilitate geo-localization. However, all of them ignore the domain difference between ground and aerial images, thus leading to inferior performance. 
Regmi and Shah \cite{Regmi_2019_ICCV} adopt a conditional GAN to generate aerial images from ground panoramas. Although it helps to bridge the cross-view domain gap, undesired scene contents are also induced in this process. 
Shi \etal \cite{shi2019optimal} propose a cross view feature transport module (CVFT) to better align ground and aerial features. However, it is hard for networks to learn geometric and feature response correspondences simultaneously. 
In contrast, our polar transform explicitly reduces the projected geometry difference between ground and aerial images, and thus eases the burden of networks.
As is clear in Table \ref{tab: compare_stoa_CVUSA} and Table \ref{tab: compare_stoa_CVACT}, our method significantly outperforms the state-of-the-art methods by a large margin.

\noindent\textbf{Fine-grained localization:}
We also compare our method with state-of-the-art methods on the CVACT\_test dataset. This dataset provides fine-grained geo-tagged aerial images that densely cover a city, and the localization performance is measured in terms of distance (meters). Specifically, a ground image is considered as successfully localized if one of the retrieved top $K$ aerial images is within 5 meters of the ground truth location of the query ground image.
Following the evaluation protocol in Liu \& Li \cite{Liu_2019_CVPR}, we plot the percentage of correctly localized ground images (recall) at different values of $K$ in Figure \ref{fig: recall_CVACT_test}.
% , and the percentage of correctly localized ground images are recorded. Following the evaluation protocol in \cite{Liu_2019_CVPR}, we plot the recall curves for different values of $K$ in Figure \ref{fig: recall_CVACT_test}.
Our method achieves superior results compared to state-of-the-art on this extremely challenging test set. 
% Figure \ref{fig:ACT_test_local_examp} show some qualitative localization results on this test set.

\begin{table*}[tbp]
\small
\setlength{\tabcolsep}{1pt}
\centering
\vspace{-2mm}
\caption{\small Comparison of recall rates for localizing ground images with unknown orientation and varying FoVs.}
\begin{tabular}{c|c|c c c c| c c c c| c c c c| c c c c}
\toprule
 \multirow{2}{*}{Dataset} & \multirow{2}{*}{Comparison Algorithms} & \multicolumn{4}{c|}{FoV=$360^{\circ}$}     & \multicolumn{4}{c}{FoV=$180^{\circ}$}     & \multicolumn{4}{c}{FoV=$90^{\circ}$}   & \multicolumn{4}{c}{FoV=$70^{\circ}$}  \\ \cline{3-18} 
                      &                                       & r@1   & r@5   & r@10  & r@1\%  & r@1   & r@5   & r@10  & r@1\%   & r@1   & r@5   & r@10  & r@1\%  & r@1   & r@5   & r@10  & r@1\%\\\hline \hline
 \multirow{3}{*}{CVUSA} 
%  & Zhai \etal \cite{zhai2017predicting} & 3.10 & 10.38 & 16.91 & 54.30  & 1.63 & 6.60 & 11.37 & 42.95 & 1.06 & 3.62 & 6.19 & 29.67  & 0.65 & 2.78 & 5.22 & 27.14   \\  
                        & CVM-NET  \cite{Hu_2018_CVPR}         & 16.25 & 38.86 & 49.41 & 88.11  & 7.38  & 22.51 & 32.63 & 75.38 & 2.76  & 10.11 & 16.74 & 55.49  & 2.62  & 9.30  & 15.06 & 21.77   \\  
                            %   & Liu \& Li \cite{Liu_2019_CVPR} & 38.83 & 64.63 & 75.02 & 95.34  & 22.20 & 44.91 & 55.98 & 88.50 & 4.04  & 11.31 & 16.94 & 49.09  & 2.72  & 8.81  & 13.67 & 43.54   \\ 
                              & CVFT \cite{shi2019optimal}     & 23.38 & 44.42 & 55.20 & 86.64  & 8.10 & 24.25 & 34.47 & 75.15  & 4.80   & 14.84 & 23.18 & 61.23  & 3.79 & 12.44 & 19.33 & 55.56    \\ 
                              & Ours  & \textbf{78.11} & \textbf{89.46} & \textbf{92.90} & \textbf{98.50} & \textbf{48.53} & \textbf{68.47} & \textbf{75.63} & \textbf{93.02}   & \textbf{16.19} & \textbf{31.44} & \textbf{39.85} & \textbf{71.13} & \textbf{8.78} & \textbf{19.90} & \textbf{27.30} & \textbf{61.20}     \\ \hline
\multirow{3}{*}{CVACT\_val}  
% & Zhai \etal \cite{zhai2017predicting} & 2.41 & 7.97 & 12.76 & 43.71  & 1.40 & 5.53 & 9.38 & 36.95 & 0.73 & 2.48 & 4.47 & 22.69  & 0.62 & 2.23 & 4.00 & 21.61   \\
                              & CVM-NET  \cite{Hu_2018_CVPR}   & 13.09 & 33.85 & 45.69 & 81.80  & 3.94  & 13.69 & 21.23 & 59.22  & 1.47 & 5.70  & 9.64  & 38.05  & 1.24 & 4.98 & 8.42 & 34.74  \\ 
                            %   & Liu \& Li \cite{Liu_2019_CVPR} & 48.65 & 69.88 & 77.34 & 93.07  & 21.53 & 44.00 & 55.03 & 85.51  & 7.26 & 19.88 & 28.34 & 64.61  & 5.66 & 16.33 & 23.74 & \textbf{60.37}    \\ 
                              & CVFT \cite{shi2019optimal}     & 26.79 & 46.89 & 55.09 & 81.03  & 7.13 & 18.47 & 26.83 & 63.87   & 1.85 & 6.28 & 10.54 & 39.25   & 1.49 & 5.13 & 8.19 & 34.59    \\ 
                             & Ours  & \textbf{72.91}  & \textbf{85.70}     & \textbf{88.88}     & \textbf{95.28} & \textbf{49.12}  & \textbf{67.83}     & \textbf{74.18}     & \textbf{89.93}  & \textbf{18.11}  & \textbf{33.34}     & \textbf{40.94}     & \textbf{68.65} & \textbf{8.29}  & \textbf{20.72}     & \textbf{27.13}     & \textbf{57.08} \\  \bottomrule

\end{tabular}
\label{tab: compare_stoa_unknown_orien_limited_FoV}
\end{table*}

\subsection{Localizing With Unknown Orientation and Limited FoV}
In this section, we test the performance of our algorithm and other methods, including CVM-NET \cite{Hu_2018_CVPR} and CVFT \cite{shi2019optimal}, on the CVUSA and CVACT\_val datasets in a more realistic localization scenario, where the ground images do not have a known orientation and have a limited FoV. Recall that Liu \& Li \cite{Liu_2019_CVPR} require orientation information as an input, so we cannot compare with this method. 

% \medskip
\noindent\textbf{Location estimation:}
Since existing methods are only designed to estimate the location of ground images, we only evaluate their location recall performance.
In order to evaluate the impact of orientation misalignments and limited FoVs on localization performance, we randomly shift and crop the ground panoramas along the azimuth direction for the CVUSA and CVACT\_val datasets. In this manner, we mimic the procedure of localizing images with limited FoV and unknown orientation.
The first results column in Table \ref{tab: compare_stoa_unknown_orien_limited_FoV} demonstrates the performance of localizing panoramas with unknown orientation. It is clear that our method significantly outperforms all the comparison algorithms, obtaining a $2.34\times$ improvement on CVUSA and a $2.72\times$ improvement on CVACT in terms of r@1.
We also conduct comparisons with the other methods on ground images with FoVs of $180^{\circ}$ (fish-eye camera), $90^{\circ}$ (wide-angle camera) and $70^{\circ}$ (general phone camera) respectively in Table \ref{tab: compare_stoa_unknown_orien_limited_FoV}. Note that the orientation is also unknown.
As illustrated in Figure \ref{fig: visual_orien_70}, as the FoV of the ground image decreases, the image become less discriminative. This increases the difficulty of geo-localization especially when the orientation is unknown. 
As indicated in the second, third and fourth results column of Table \ref{tab: compare_stoa_unknown_orien_limited_FoV}, our method, benefiting from its DSM module, significantly reduces the ambiguity caused by unknown orientations and measures feature similarity more accurately, achieving better performance than the state-of-the-art.
% The recall curves from top 1 to top 1\% of the comparison algorithms on ground images with different FoVs are provided in the supplementary material.

\begin{figure}[!t] 
\centering
\hspace{-2.6mm}
\subfigure[ Query]{
\tiny
\vspace{-2mm}
\begin{minipage}{0.385\linewidth}
\vspace{-2mm}
\centering
\vspace{-2mm}
\includegraphics[width=\textwidth, height=0.33\textwidth]{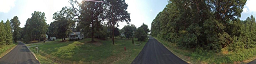}
\centerline{\tiny FoV=$360^{\circ}$, Azimuth=$-32.344^\circ$ }
% \vspace{-2mm}
\includegraphics[width=0.5\textwidth, height=0.33\textwidth]{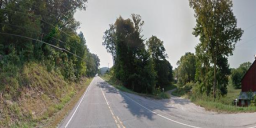}
\centerline{\tiny FoV=$180^{\circ}$, Azimuth=$128.672^\circ$ }
% \vspace{-2mm}
\includegraphics[width=0.25\textwidth, height=0.33\textwidth]{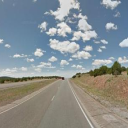}
\centerline{\tiny FoV=$90^{\circ}$, Azimuth=$-158.906^\circ$ }
% \vspace{-2mm}
\includegraphics[width=0.2\textwidth, height=0.33\textwidth]{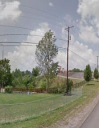}
\centerline{\tiny FoV=$70^{\circ}$, Azimuth=$-115.469^\circ$ }
\vspace{-1mm}
\end{minipage}
}
\hspace{-2mm}
\subfigure[ Top-1]{
\tiny
\begin{minipage}{0.13\linewidth}
\vspace{-2mm}
\centering
\vspace{-2mm}
\includegraphics[width=\textwidth]{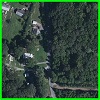}
\centerline{\tiny $-33.750^\circ$}
% \vspace{-2mm}
\includegraphics[width=\textwidth]{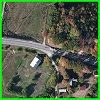}
\centerline{\tiny $129.375^\circ$ }
% \vspace{-2mm}
\includegraphics[width=\textwidth]{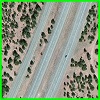}
\centerline{\tiny $-157.50^\circ$ }
% \vspace{-2mm}
\includegraphics[width=\textwidth]{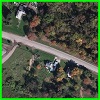}
\centerline{\tiny $-116.875^\circ$ }
\vspace{-1mm}
\end{minipage}
}
\hspace{-2mm}
\subfigure[ Top-2]{
\tiny
\begin{minipage}{0.13\linewidth}
\vspace{-2mm}
\centering
\vspace{-2mm}
\includegraphics[width=\textwidth]{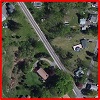}
\centerline{\tiny $-61.875^\circ$ }
% \vspace{-2mm}
\includegraphics[width=\textwidth]{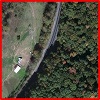}
\centerline{\tiny $-129.375^\circ$ }
% \vspace{-2mm}
\includegraphics[width=\textwidth]{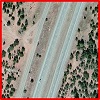}
\centerline{\tiny $-163.125^\circ$ }
% \vspace{-2mm}
\includegraphics[width=\textwidth]{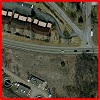}
\centerline{\tiny $57.500^\circ$ }
\vspace{-1mm}
\end{minipage}
}
\hspace{-2mm}
\subfigure[ Top-3]{
\tiny
\begin{minipage}{0.13\linewidth}
\vspace{-2mm}
\centering
\vspace{-2mm}
\includegraphics[width=\textwidth]{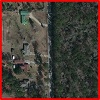}
\centerline{\tiny $-28.125^\circ$ }
% \vspace{-2mm}
\includegraphics[width=\textwidth]{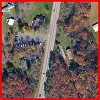}
\centerline{\tiny $-146.250^\circ$}
% \vspace{-2mm}
\includegraphics[width=\textwidth]{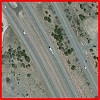}
\centerline{\tiny $-28.125^\circ$}
% \vspace{-2mm}
\includegraphics[width=\textwidth]{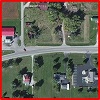}
\centerline{\tiny $-128.125^\circ$ }
\vspace{-1mm}
\end{minipage}
}
\hspace{-2mm}
\subfigure[ Top-4]{
\tiny
\begin{minipage}{0.13\linewidth}
\vspace{-2mm}
\centering
\vspace{-2mm}
\includegraphics[width=\textwidth]{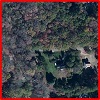}
\centerline{\tiny $-123.750^\circ$ }
% \vspace{-2mm}
\includegraphics[width=\textwidth]{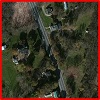}
\centerline{\tiny $-180.000^\circ$ }
% \vspace{-2mm}
\includegraphics[width=\textwidth]{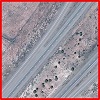}
\centerline{\tiny $39.375^\circ$}
% \vspace{-2mm}
\includegraphics[width=\textwidth]{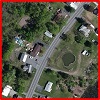}
\centerline{\tiny $175.635^\circ$ }
\vspace{-1mm}
\end{minipage}
}
\hspace{-2mm}
\\
\vspace{-3mm}
\caption{\small{
Visualization of joint location and orientation estimation results by our method on the CVUSA dataset. The FoV and ground truth azimuth angle are presented under each query image. The angle under each aerial image is the estimated relative orientation of the query image with respect to this aerial image. 
% From left to right: ground-level query image and the top 1-5 retrieved aerial candidates. 
Green and red borders indicate correct and wrong retrieved results, respectively.
}}
\label{fig: CVUSA_unkown_orien_FOV_localization_example}
\end{figure}

% \medskip
\noindent\textbf{Orientation estimation:}
As previously mentioned, the experiments of orientation estimation is conducted on ground images which are correctly localized in terms of top-1 retrieved candidates. 
The first row in Table~\ref{tab: orien_acc_top1} presents the orientation prediction accuracy of ground images with different FoVs. 
As indicated in the table, the orientation of almost all ground images with $360^\circ$ and $180^{\circ}$ FoV is predicted correctly, demonstrating the effectiveness of our DSM module for estimating the orientation of ground images.
It is also clear that the matching ambiguity increases as the FoV decreases.
Considering that scene contents in an aerial image might be very similar in multiple directions, the orientation estimation can be inaccurate while the estimated location is correct. For instance, a person standing on a road is able to localize their position but will find it difficult to determine their orientation if the view is similar along the road in both directions. 
We provide an example of this in the supplementary material.
Therefore, even when our method estimates orientation inaccurately, it is still possible to localize the position correctly using our DSM module. 
We also report the median value of the errors (in degrees) between the estimated and ground truth orientation in the second row of Table \ref{tab: orien_acc_top1}. 
The estimated errors are very small with respect to the FoV of the image, and so will not negatively affect the localization performance.
Figure \ref{fig:visual_orien} shows the estimated orientation of ground images with $360^{\circ}$ and $70^{\circ}$ FoVs, and Figure \ref{fig: CVUSA_unkown_orien_FOV_localization_example} presents some qualitative examples on joint location and orientation estimation.
More visualization results on orientation estimation are provided in the supplementary material.

\begin{figure}[!t]
    \centering
    \vspace{-2mm}
    \subfigure[FoV=$360^{\circ}$]{
    \vspace{-2mm}
    \begin{minipage}[b]{\linewidth}
    \begin{minipage}[b]{0.136\linewidth}
    \centering
    \includegraphics[width=\linewidth]{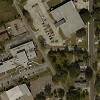}
    \end{minipage}
    \hspace{-1mm}
    \begin{minipage}[b]{0.42\linewidth}
    \centering
    \includegraphics[width=\linewidth,height=0.3333\linewidth]{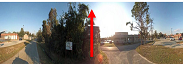}
    \end{minipage}
    \hspace{-1.5mm}
    \begin{minipage}[b]{0.42\linewidth}
    \centering
    \scalebox{1.0}[1.0]{\includegraphics[trim={2mm 0mm 0mm 3mm}, clip, width=\linewidth,height=0.3333\linewidth]{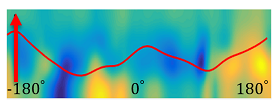}}
    \end{minipage}\\
     \begin{minipage}[b]{0.136\linewidth}
    \centering
    \includegraphics[width=\textwidth]{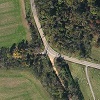}
    \end{minipage}
    \hspace{-1mm}
    \begin{minipage}[b]{0.42\linewidth}
    \centering
    \includegraphics[width=\linewidth,height=0.3333\linewidth]{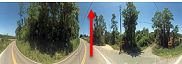}
    \end{minipage}
    \hspace{-1.5mm}
    \begin{minipage}[b]{0.42\linewidth}
    \centering
    \scalebox{1.0}[1.0]{\includegraphics[trim={0mm 0mm 0mm 3mm}, clip, width=\linewidth,height=0.3333\linewidth]{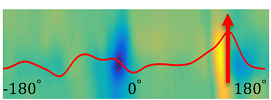}}
    \end{minipage}\\
    \vspace{-1em}
     \end{minipage}
     \label{fig: visual_orien_360}
     }\\
     \subfigure[FoV=$70^{\circ}$]{
     \vspace{-2mm}
    \begin{minipage}[b]{\linewidth}
    \begin{minipage}[b]{0.136\linewidth}
    \centering
    \includegraphics[width=\linewidth]{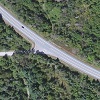}
    \end{minipage}
    \hspace{-1mm}
    \begin{minipage}[b]{0.42\linewidth}
    \centering
    \includegraphics[height=0.3333\linewidth]{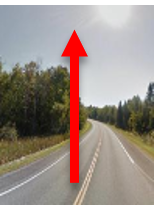}
    \end{minipage}
    \hspace{-1mm}
    \begin{minipage}[b]{0.42\linewidth}
    \centering
    \scalebox{1.0}[1.0]{\includegraphics[trim={0mm 0mm 0mm 3mm}, clip, width=\linewidth,height=0.3333\linewidth]{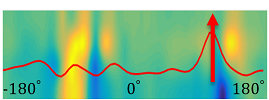}}
    \end{minipage}\\
     \begin{minipage}[b]{0.136\linewidth}
    \centering
    \includegraphics[width=\textwidth]{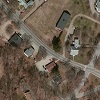}
    \end{minipage}
    \hspace{-1mm}
    \begin{minipage}[b]{0.42\linewidth}
    \centering
    \includegraphics[height=0.3333\linewidth]{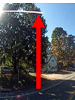}
    \end{minipage}
    \hspace{-1.5mm}
    \begin{minipage}[b]{0.42\linewidth}
    \centering
    \scalebox{1.0}[1.0]{\includegraphics[trim={0mm 0mm 0mm 3mm}, clip, width=\linewidth,height=0.3333\linewidth]{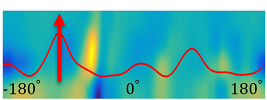}}
    \end{minipage}\\
    \vspace{-1em}
     \end{minipage}
     \label{fig: visual_orien_70}}\\
     \vspace{-5mm}
    \caption{\small Visualization of estimated orientation for ground images with FoV = $360^{\circ}$ and $70^{\circ}$. In each of the subfigures, the aerial images are on the left and the ground images are in the middle. We visualize the polar transformed-aerial features and the correlation results (red curves) in the right column. The positions of the correlation maxima in the curves corresponds to the orientation of the ground images.}
    \label{fig:visual_orien}
\end{figure}

\begin{table}[t]
\small
\setlength{\tabcolsep}{1.5 pt}
\centering
\caption{\small Orientation prediction performance on correctly localized ground images.
}
\vspace{-2mm}
\label{tab: orien_acc_top1}
\begin{tabular}{c |c c c c|c c c c}
\toprule
Dataset     & \multicolumn{4}{c|}{CVUSA}                                        & \multicolumn{4}{c}{CVACT\_val}  \\ \hline
FoV         & $360^{\circ}$ & $180^{\circ}$ & $90^{\circ}$& $70^{\circ}$        & $360^{\circ}$ & $180^{\circ}$ & $90^{\circ}$& $70^{\circ}$   \\ \hline \hline
orien\_acc   & 99.41 & 98.54 & 76.15 & 61.67                                     & 99.84 & 99.10 & 74.51 & 55.18   \\
median\_error& 2.38 & 2.38 & 4.50 & 4.88                                     & 1.97 & 2.89 & 5.21 & 6.22    \\
\bottomrule

\end{tabular}
\end{table}

\section{Conclusion}
In this paper we have proposed an effective algorithm for image-based geo-localization, which can handle complex situation when neither location nor orientation is known.  
% Armed with this algorithm, and take a camera and a satellite map with you, you will never get lost or disoriented on hiking trails. 
Contrast to many existing methods, our algorithm recovers both location and orientation by joint cross-view image matching. 
% Key contributions of this paper include a polar-transformation to bring different domains closer and a novel Dynamic Similarity Matching module (DSM) to regress on relative orientation. 
Key components of our framework include a polar-transformation to bring different domains closer and a novel Dynamic Similarity Matching module (DSM) to regress on relative orientation. 
Benefited from the two items, our network is able to extract appropriate aerial features if the ground image is disoriented and has a limited FoV.  We obtained higher location recalls for cross-view image matching, significantly improve the state-of-the-art in multitude practical scenarios.

\section{ Acknowledgments}
This research is supported in part by the Australian Research Council (ARC) Centre of Excellence for Robotic Vision (CE140100016),  ARC-Discovery (DP 190102261) and ARC-LIEF (190100080),  as well as a research grant from Baidu on autonomous driving. 
The first author is a China Scholarship Council (CSC)-funded PhD student to ANU. We gratefully acknowledge the GPUs donated by the NVIDIA Corporation. We thank all anonymous reviewers and ACs for their constructive comments.

{\small
\bibliographystyle{ieee_fullname}
\bibliography{egbib}
}

\newpage
\onecolumn
\appendix

\section{Localization with Unknown Orientation and Limited FoV}

\subsection{Location Estimation}
\begin{figure*}[!h]
	\centering
	\scalebox{1.0}[1.0]{\includegraphics[trim={10mm 70mm 10mm 70mm}, clip, width=0.24\linewidth]{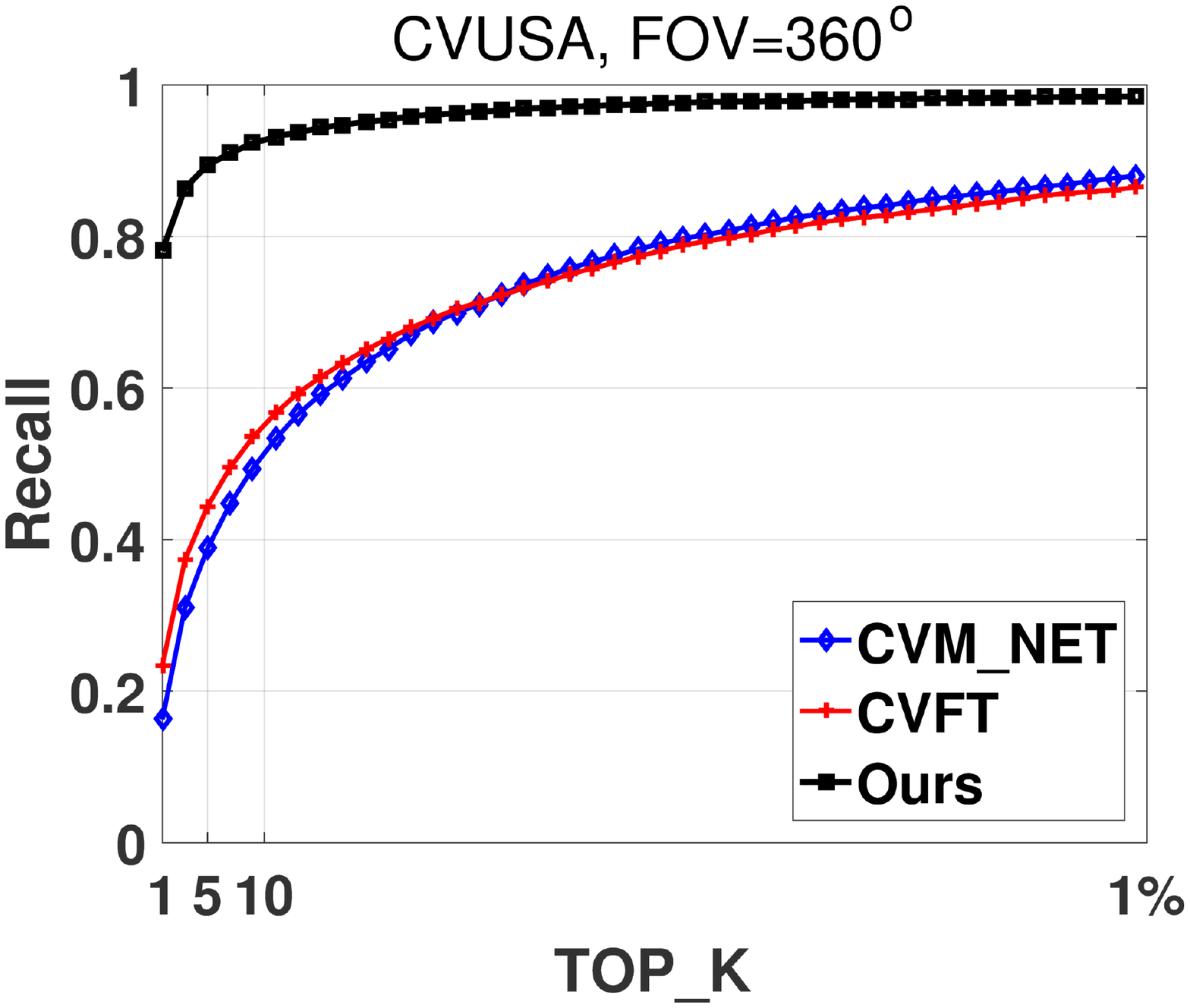}}
	\scalebox{1.0}[1.0]{\includegraphics[trim={10mm 70mm 10mm 70mm}, clip, width=0.24\linewidth]{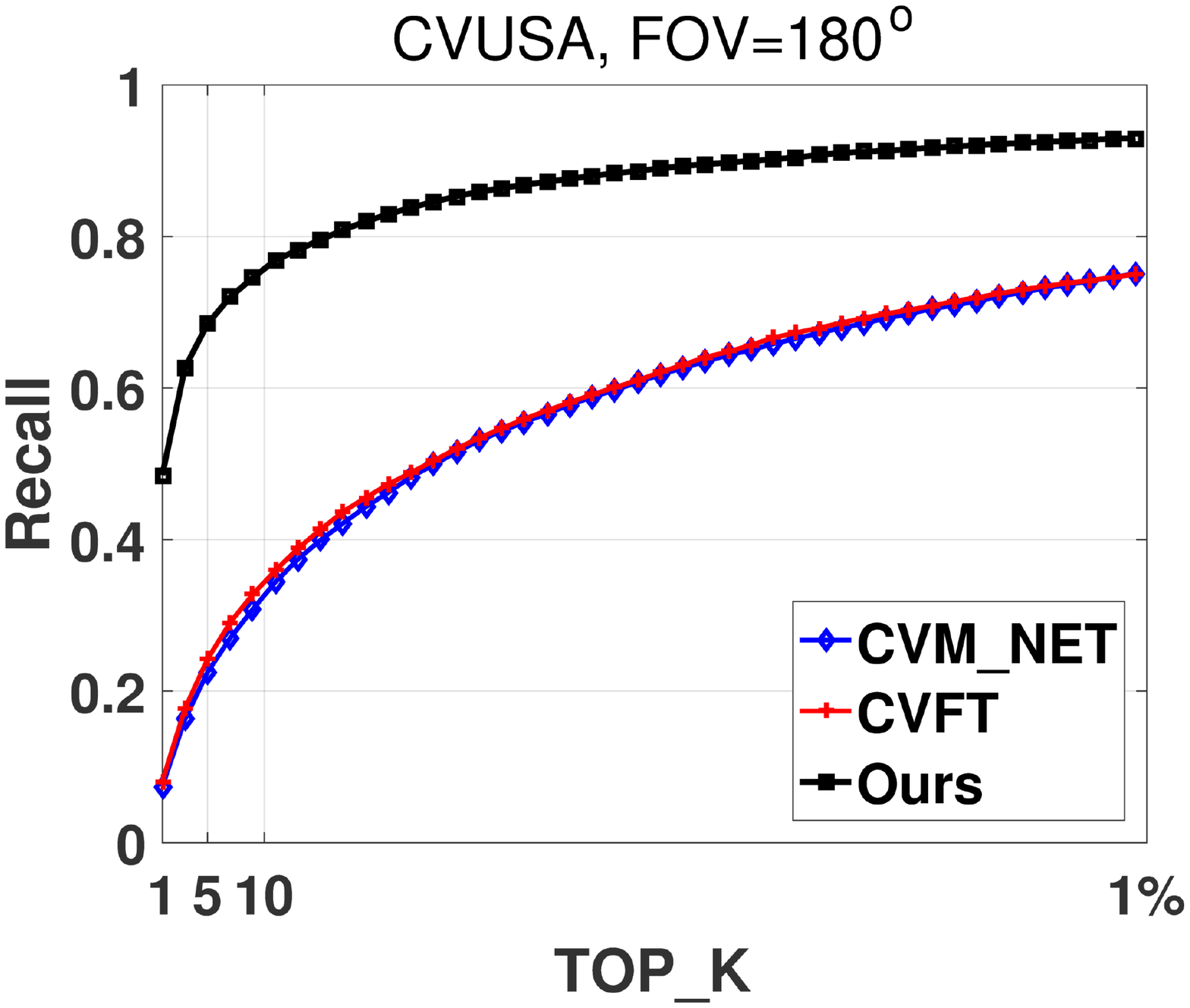}}
	\scalebox{1.0}[1.0]{\includegraphics[trim={10mm 70mm 10mm 70mm}, clip, width=0.24\linewidth]{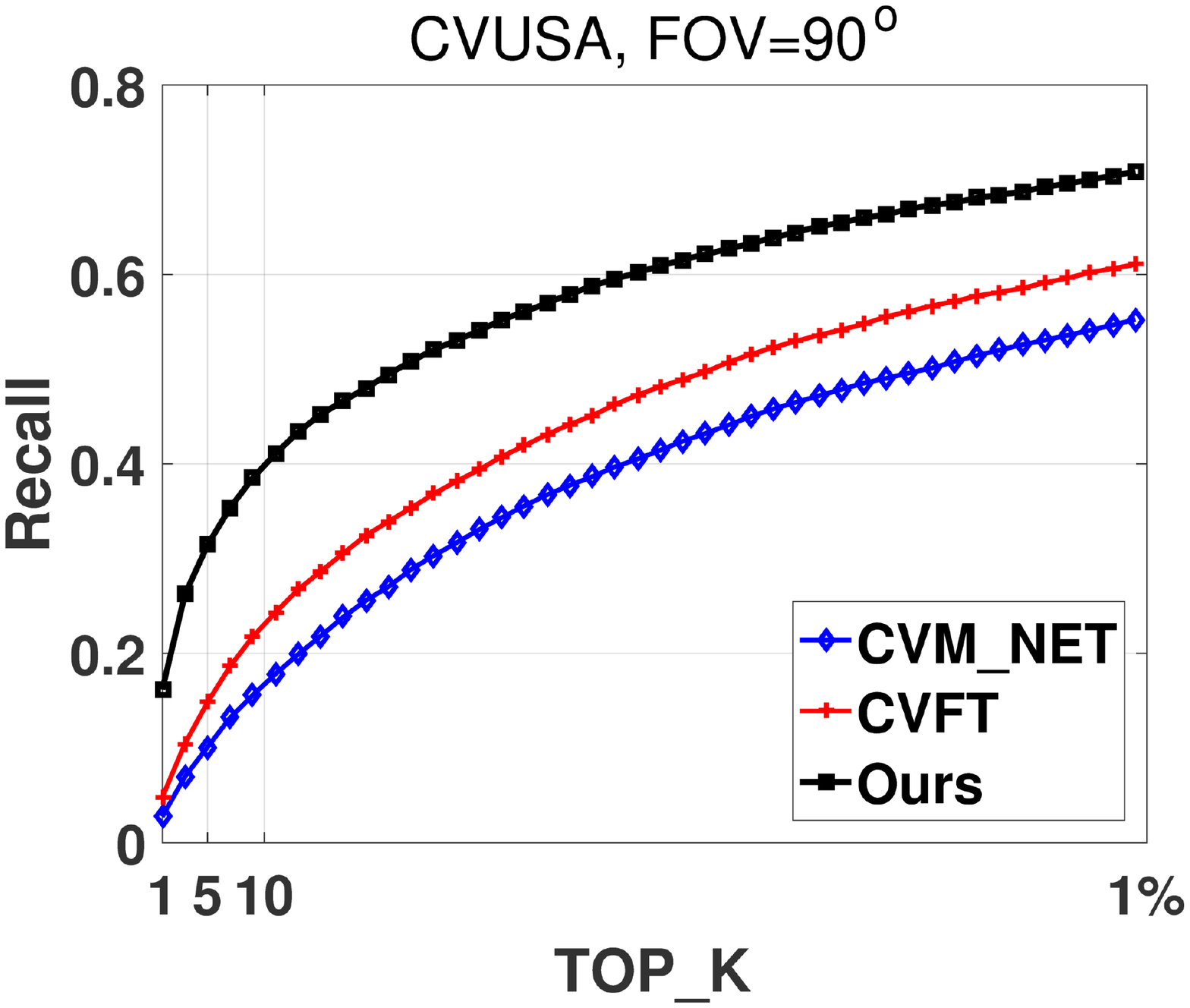}}
	\scalebox{1.0}[1.0]{\includegraphics[trim={10mm 70mm 10mm 70mm}, clip, width=0.24\linewidth]{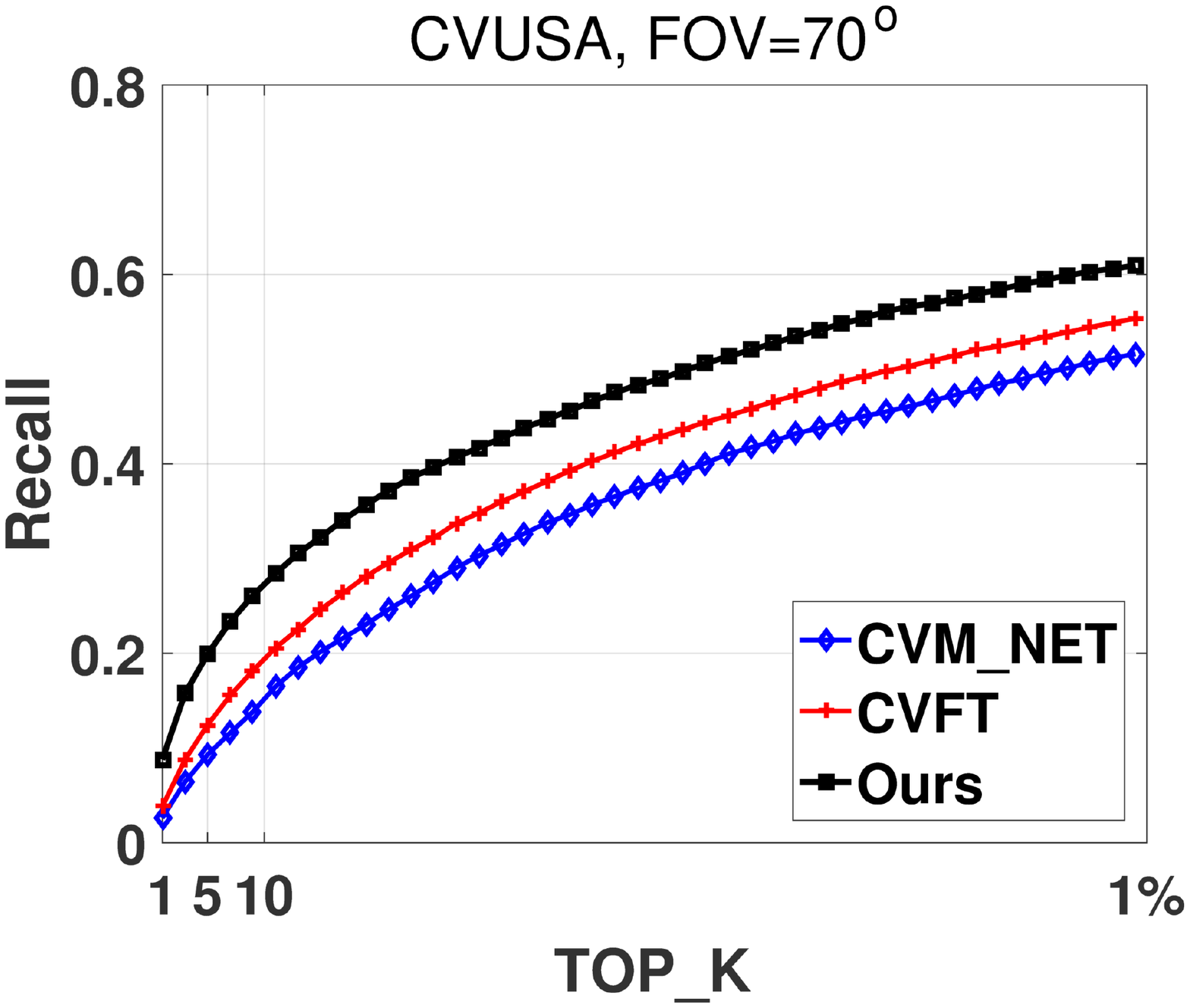}}
	\scalebox{1.0}[1.0]{\includegraphics[trim={10mm 70mm 10mm 70mm}, clip, width=0.24\linewidth]{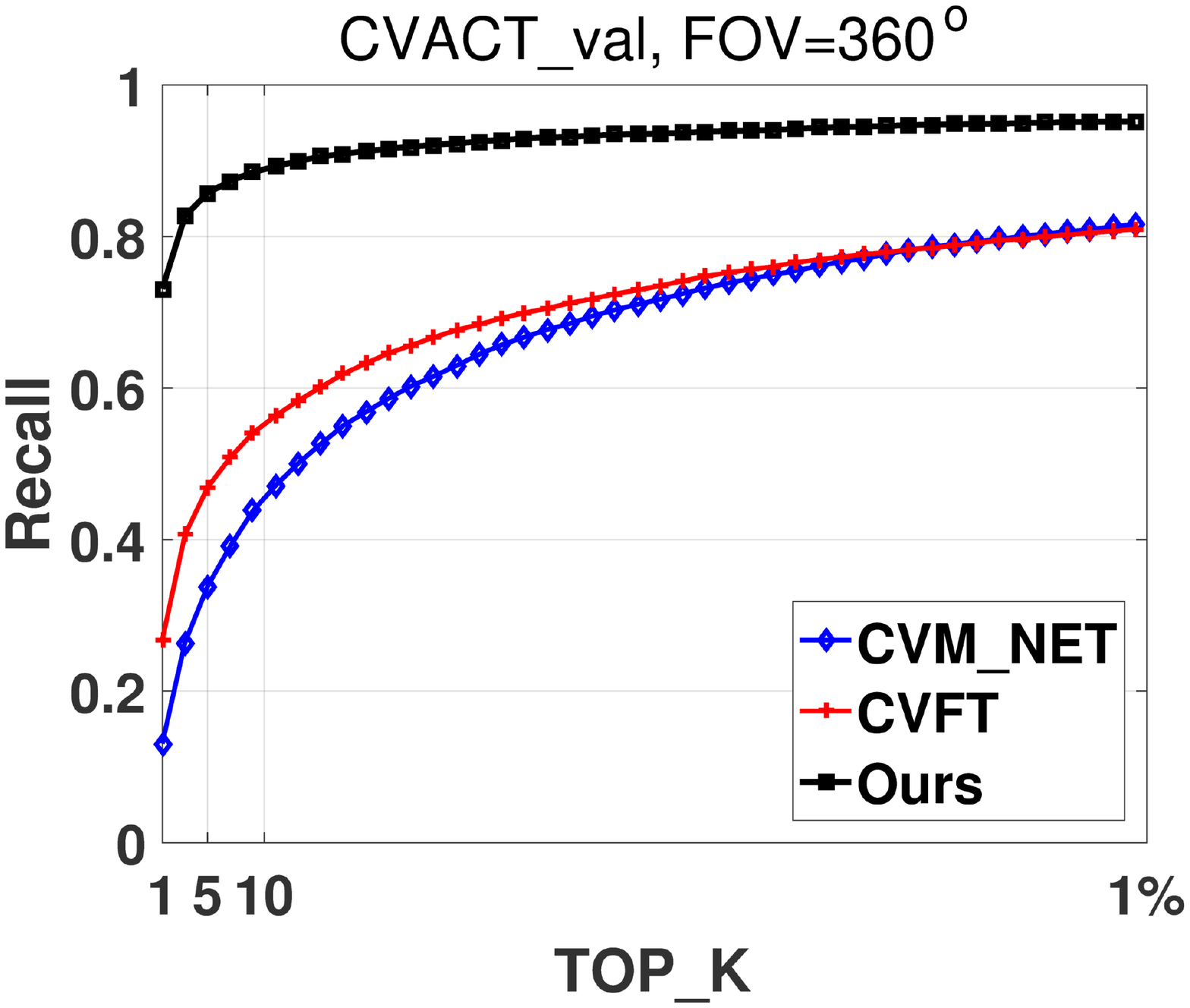}}
	\scalebox{1.0}[1.0]{\includegraphics[trim={10mm 70mm 10mm 70mm}, clip, width=0.24\linewidth]{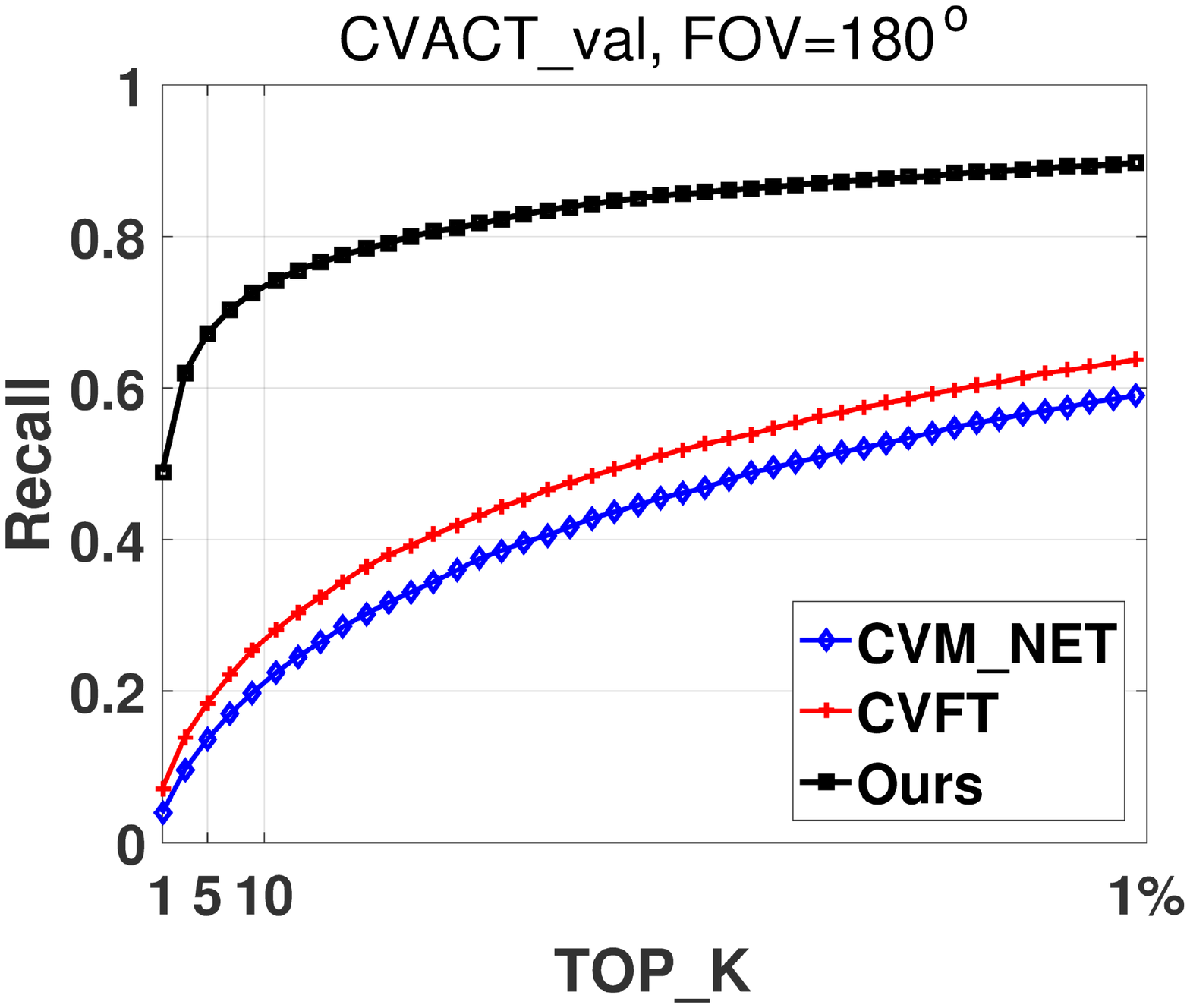}}
	\scalebox{1.0}[1.0]{\includegraphics[trim={10mm 70mm 10mm 70mm}, clip, width=0.24\linewidth]{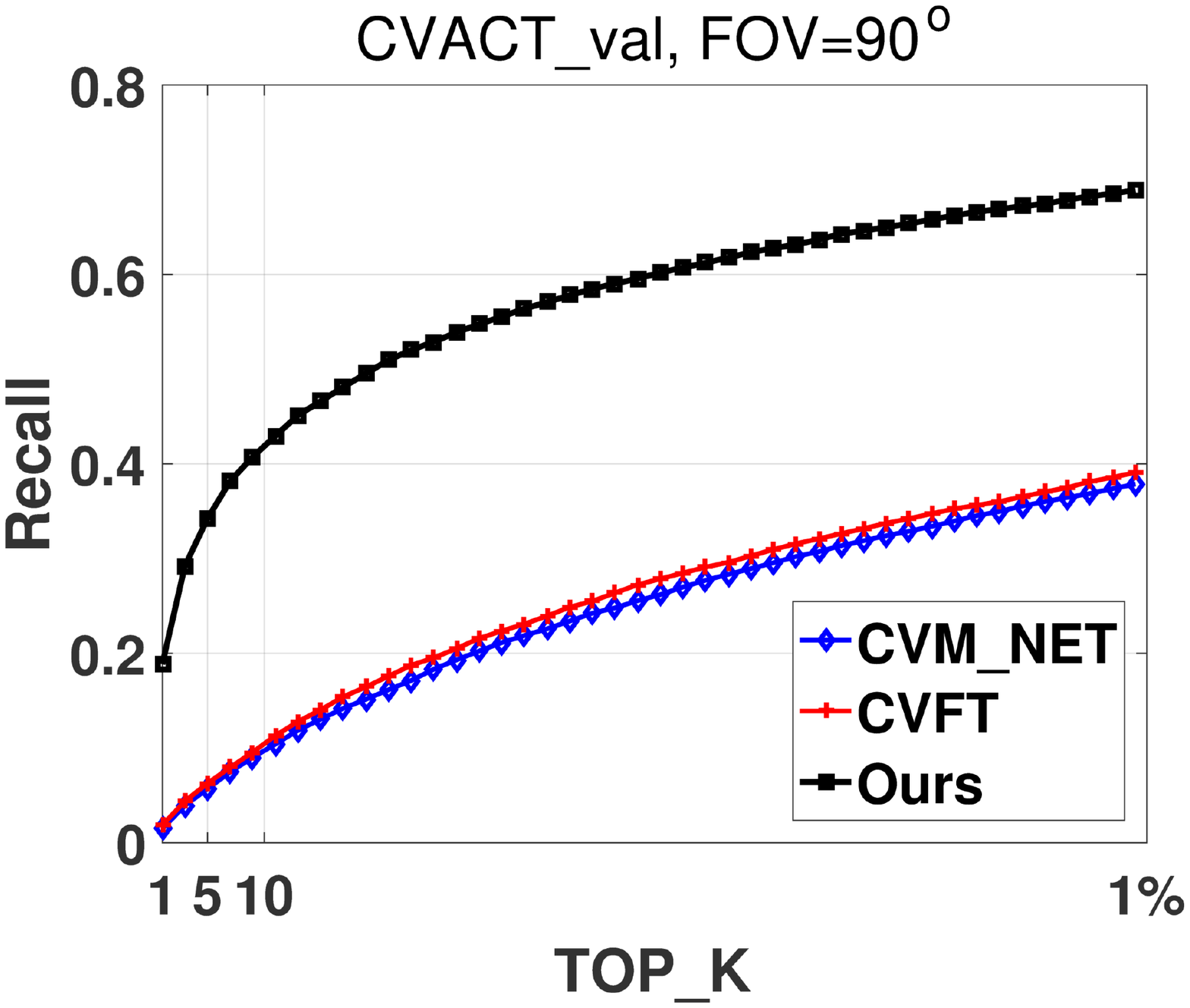}}
	\scalebox{1.0}[1.0]{\includegraphics[trim={10mm 70mm 10mm 70mm}, clip, width=0.24\linewidth]{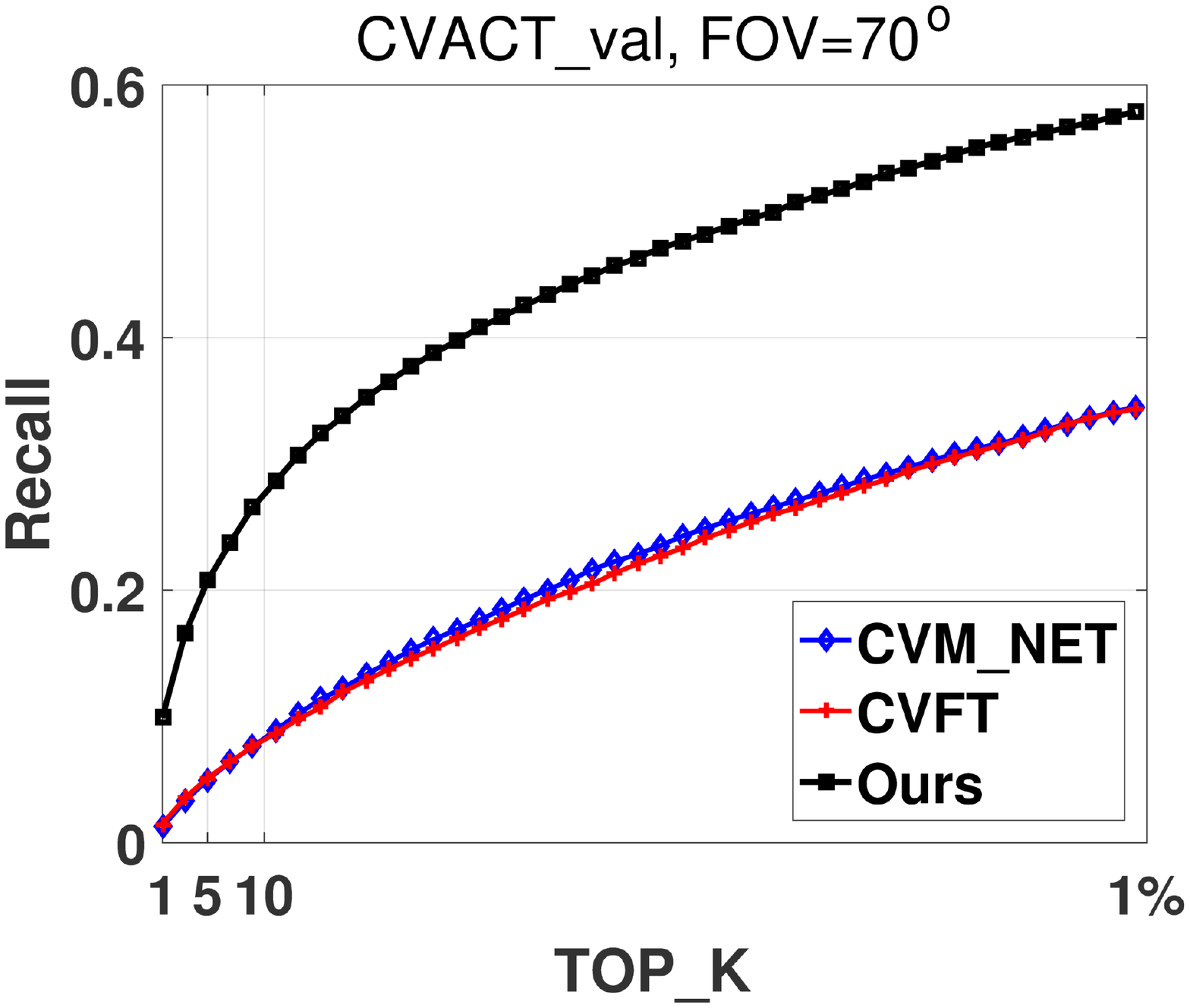}}
	
	\caption{Location estimation performance (r@$K$) of different algorithms on unknown orientation and varying FoVs.}
	\label{fig:unknown_orien_varying_FOV}
\end{figure*}

In the main paper, we report the top-1, top-5, top-10 and top-1\% recall rates of our algorithm and the state-of-the-art on localizing ground images with unknown orientation and varying FoVs. 
In this section, we present the complete r@$K$ performance in Figure \ref{fig:unknown_orien_varying_FOV}. It can be seen that our method achieves consistently better performance than the state-of-the-art algorithms in all the localization scenarios.

\begin{figure*}[!ht]
	\centering
	\scalebox{1.0}[1.0]{\includegraphics[trim={10mm 70mm 10mm 70mm}, clip, width=0.24\linewidth]{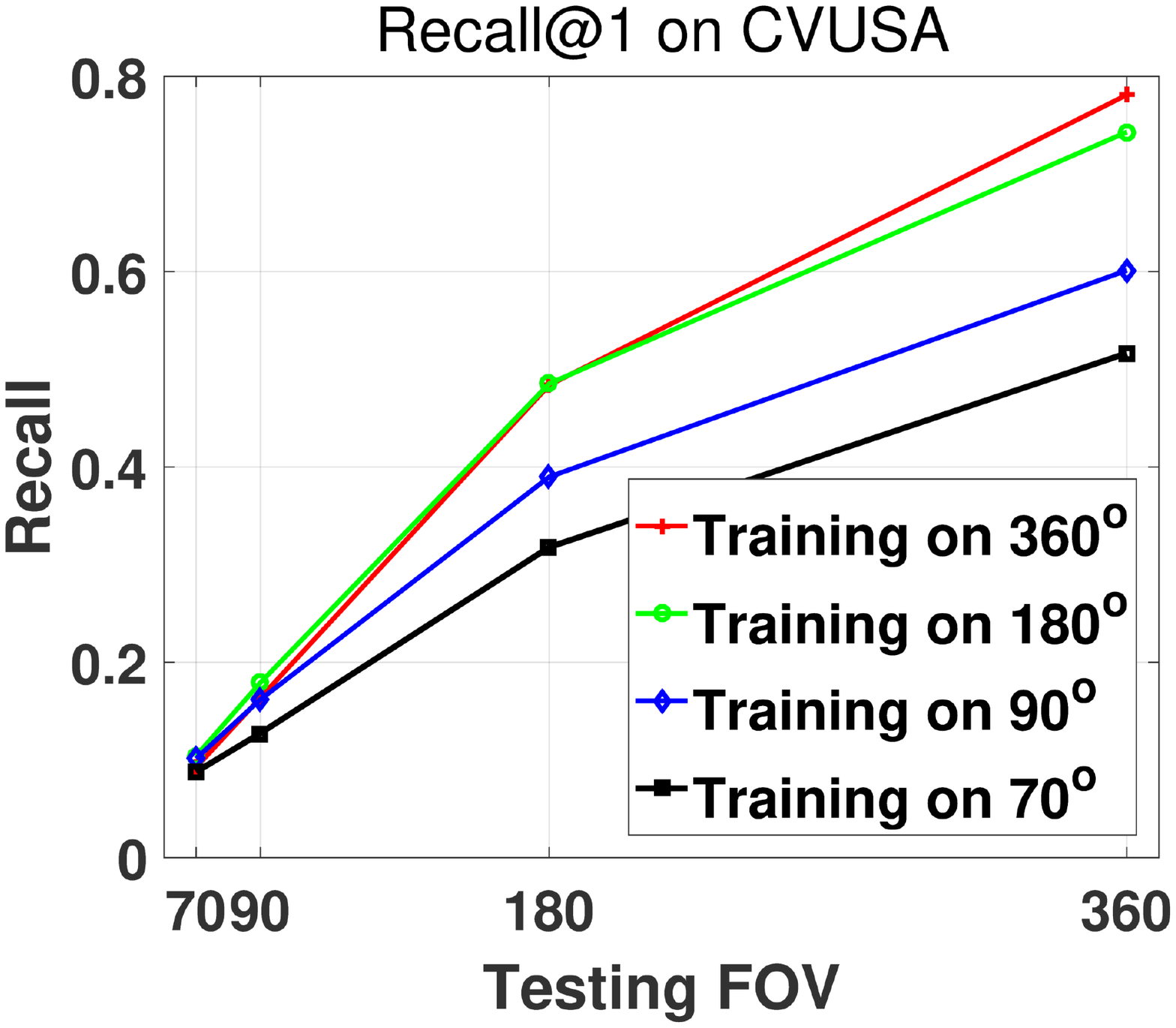}}
	\scalebox{1.0}[1.0]{\includegraphics[trim={10mm 70mm 10mm 70mm}, clip, width=0.24\linewidth]{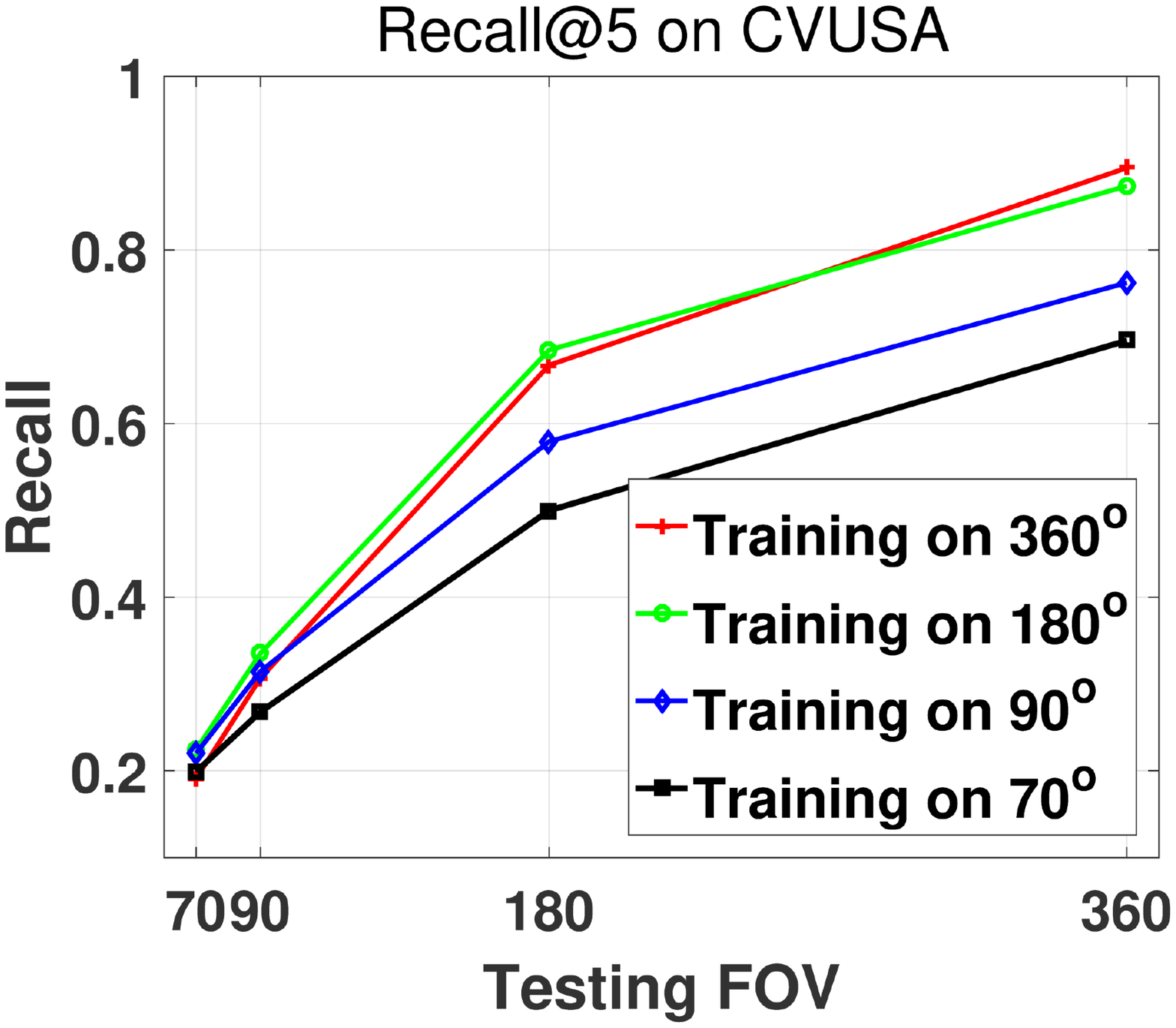}}
	\scalebox{1.0}[1.0]{\includegraphics[trim={10mm 70mm 10mm 70mm}, clip, width=0.24\linewidth]{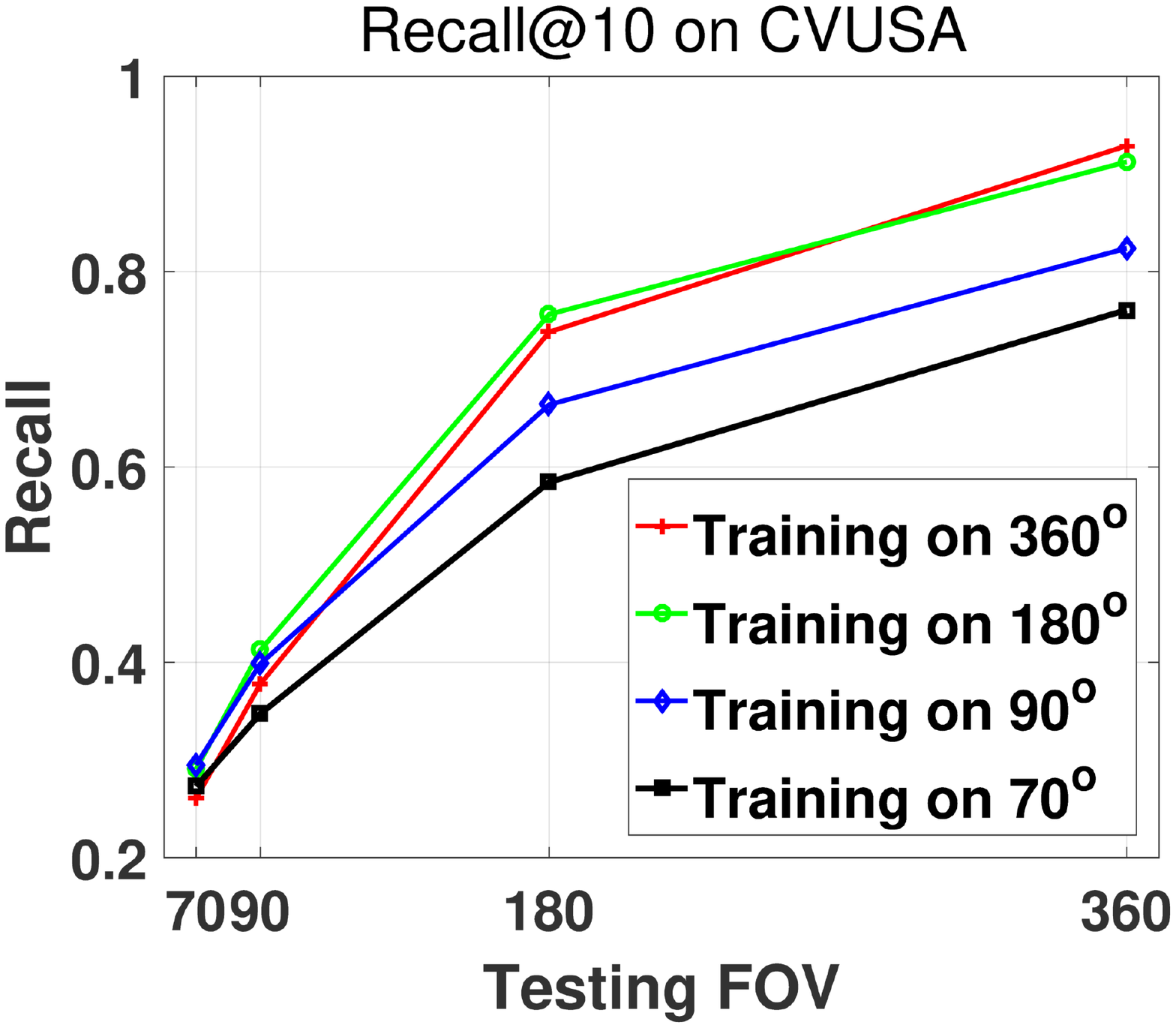}}
	\scalebox{1.0}[1.0]{\includegraphics[trim={10mm 70mm 10mm 70mm}, clip, width=0.24\linewidth]{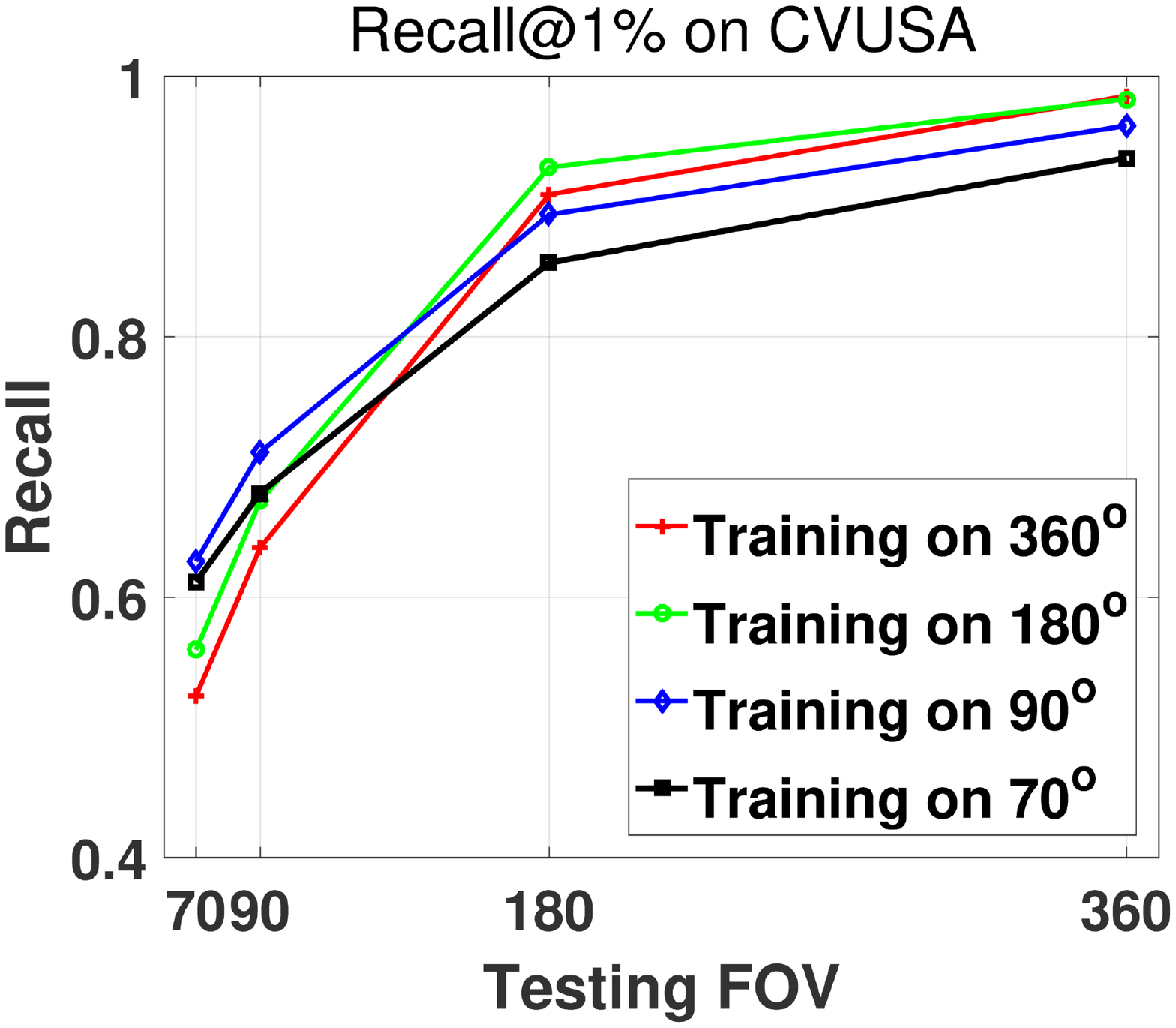}}\\
	\scalebox{1.0}[1.0]{\includegraphics[trim={10mm 70mm 10mm 70mm}, clip, width=0.24\linewidth]{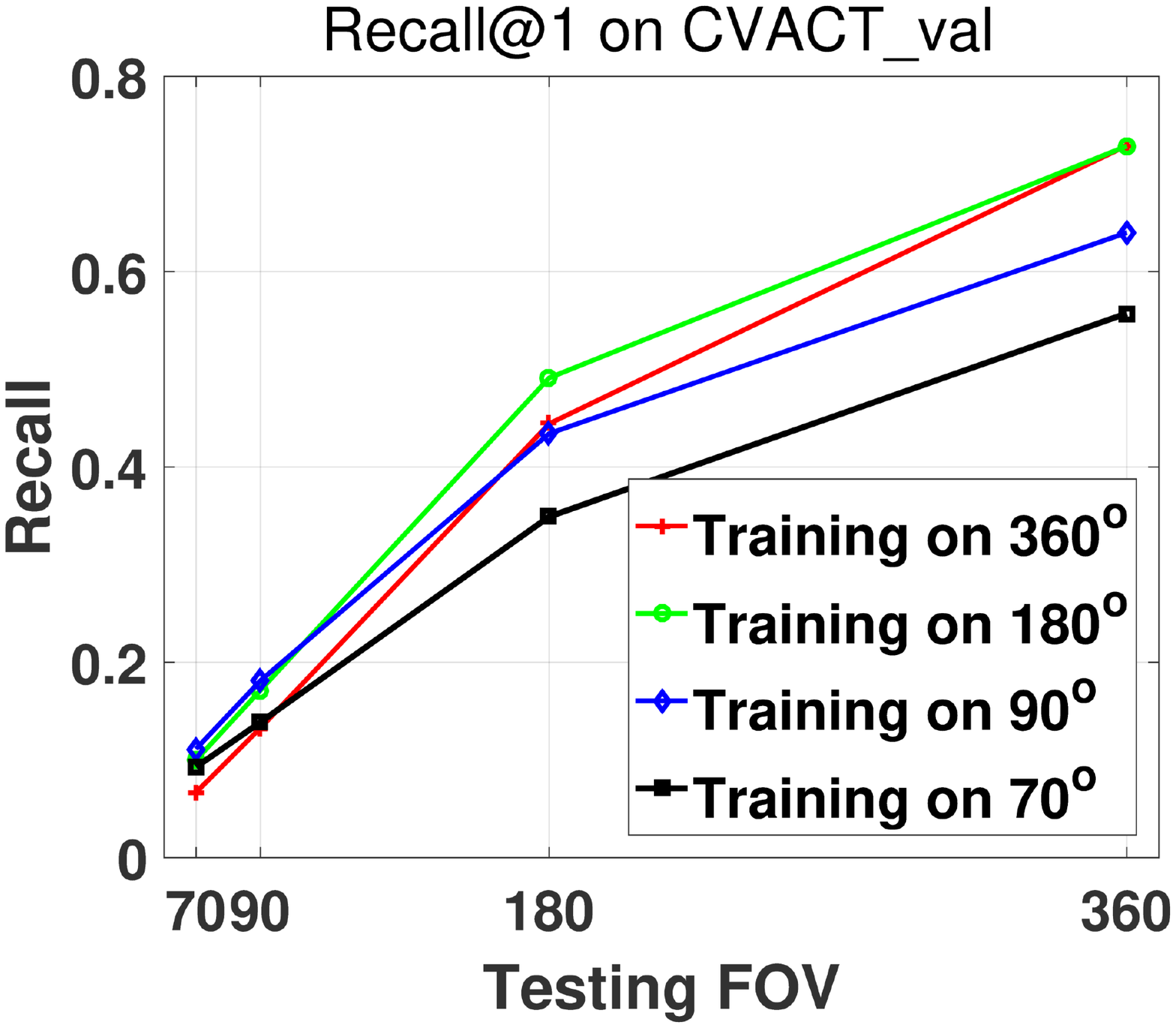}}
	\scalebox{1.0}[1.0]{\includegraphics[trim={10mm 70mm 10mm 70mm}, clip, width=0.24\linewidth]{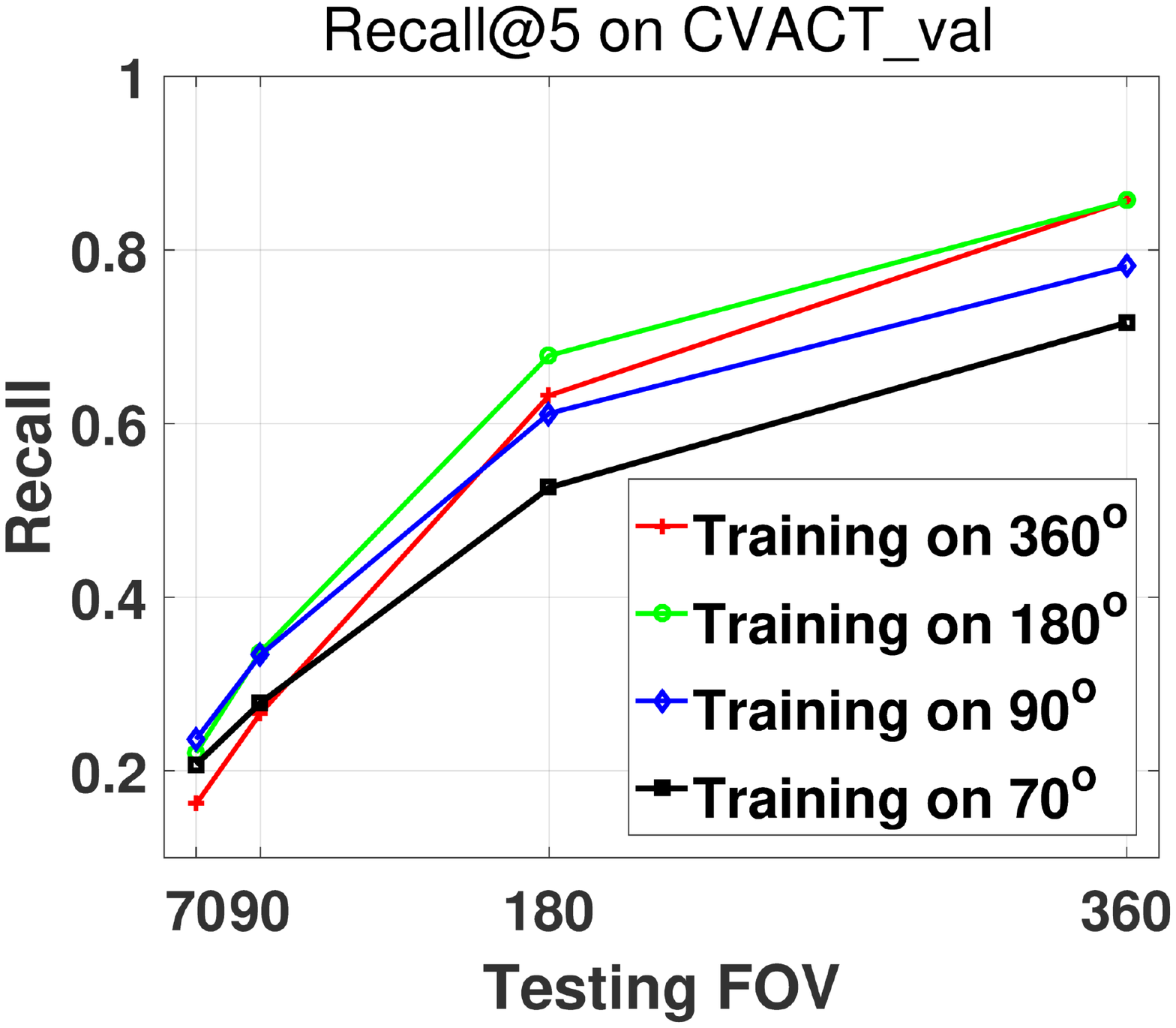}}
	\scalebox{1.0}[1.0]{\includegraphics[trim={10mm 70mm 10mm 70mm}, clip, width=0.24\linewidth]{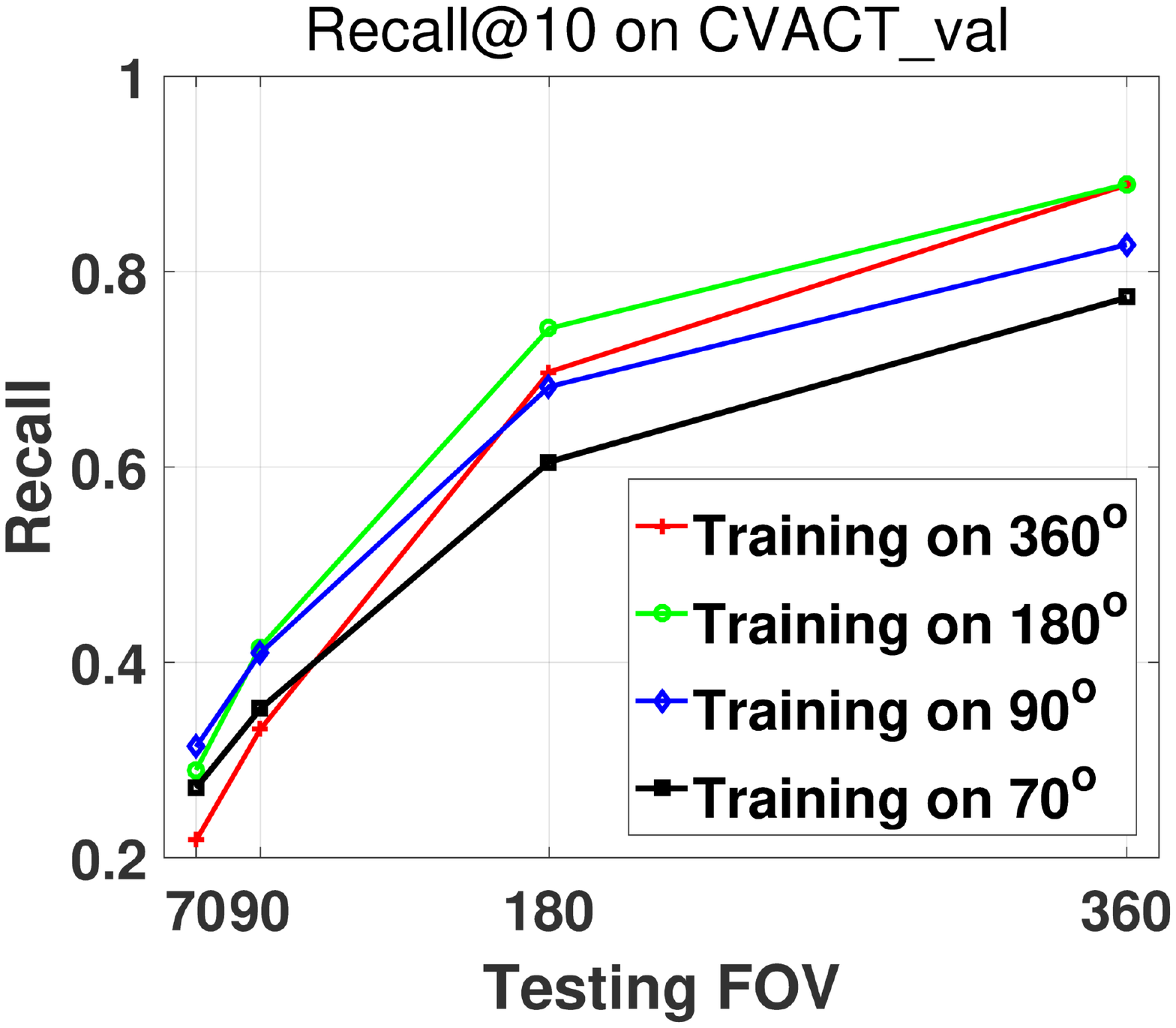}}
	\scalebox{1.0}[1.0]{\includegraphics[trim={10mm 70mm 10mm 70mm}, clip, width=0.24\linewidth]{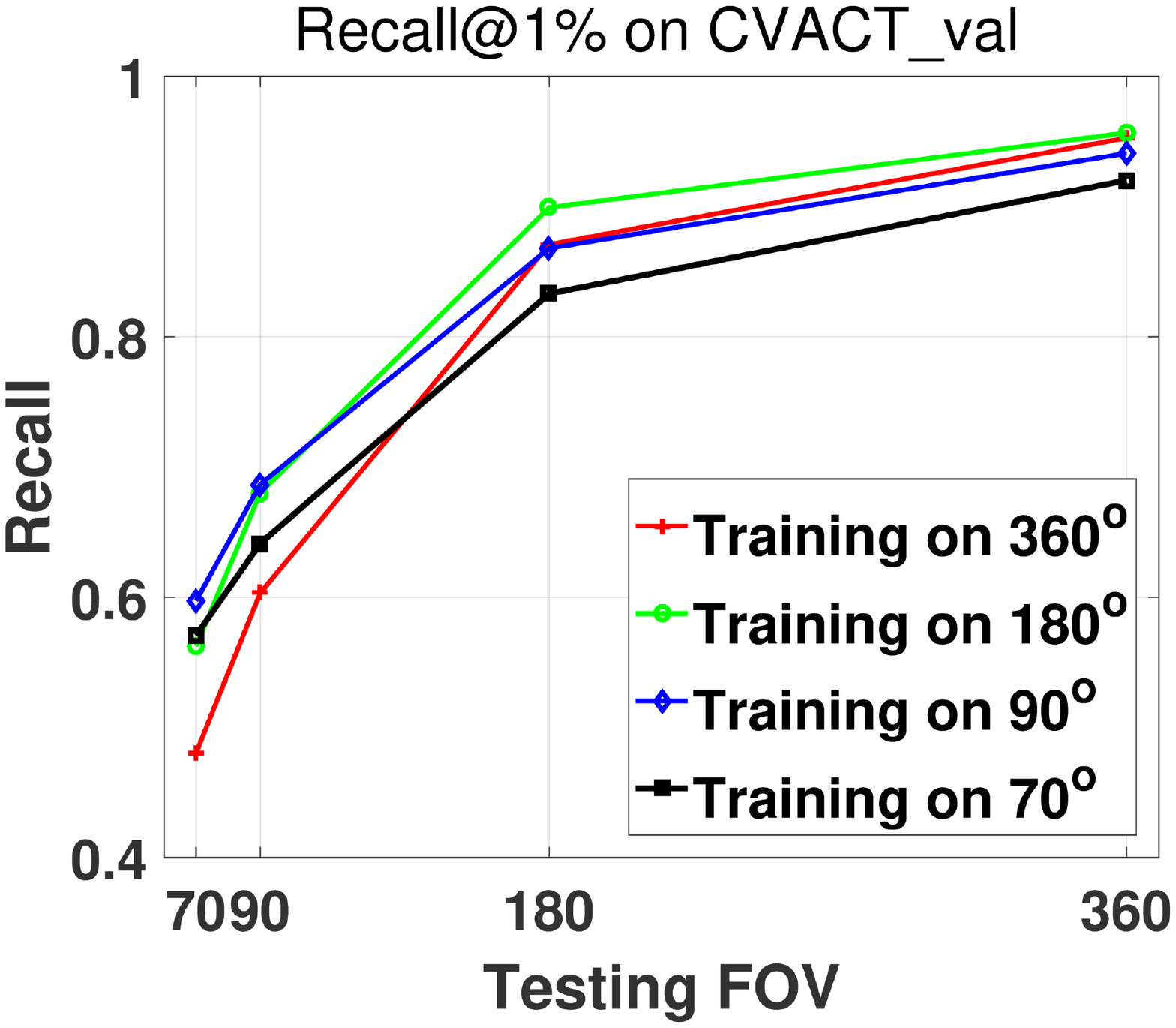}}
	\caption{Recall performance at top-1, top-5, top-10 and top-1\% of our method with different training and testing FoVs.}
	\label{fig:train_test_different_FOV}
\end{figure*}

\medskip\noindent\textbf{Training and Testing on Different FoVs:}
In real-world scenarios, a camera's FoV at inference time may not be the same as that used during training. Therefore, we also investigate the impact of using a model trained on a different FoV.
We employ a model trained on ground images with a specific FoV and test its performance on ground images with varying FoVs.
Figure \ref{fig:train_test_different_FOV} illustrates the recall curves at top-1, top-5, top-10 and top-1\% with respect to different testing FoVs, and the numerical results are presented in Table \ref{tab: train_test_different_FoV}. 
% We report the full recall curves and numerical results in the supplementary material.

It is apparent in Figure \ref{fig:train_test_different_FOV} that, as the test FoV increases, all models attain better performance. This implies that having a greater amount of scene contents reduces the matching ambiguity, and our method is able to exploit such information to achieve better localization. Furthermore, using a model trained with a FoV similar to the test image produces better results in general. Therefore, it is advisable to adopt a pretrained model with FoV similar to the test image. 

% We report the recall curves at top-5 and top-1\% of different models on different testing FoVs in the main paper. Furthermore, we provide the recall curves at top-1 and top-10 in Figure \ref{fig:train_test_different_FOV}. The numerical results at top-1, top-5, top-10 and top-1\% are presented in Table \ref{tab: train_test_different_FoV}.     

\subsection{Orientation Estimation}

\label{section: orien_estimation}

\begin{table*}[!h]
	\small
	\setlength{\tabcolsep}{2.5pt}
	\centering
	\caption{\small Numerical recall results of our method with different training and testing FoVs. }
	\label{tab: train_test_different_FoV}
	\begin{tabular}{c c|c c c c|c c c c|c c c c|c c c c}
		\toprule
		\multicolumn{2}{c|}{\multirow{2}{*}{\backslashbox{Train FoV}{Test FoV}}} & \multicolumn{4}{c|}{$360^{\circ}$}   & \multicolumn{4}{c|}{$180^{\circ}$} & \multicolumn{4}{c|}{$90^{\circ}$} & \multicolumn{4}{c}{$70^{\circ}$} \\ \cline{3-18} 
		&     & r@1   & r@5   & r@10  & r@1\%    & r@1   & r@5   & r@10  & r@1\%   & r@1   & r@5   & r@10   & r@1\%   & r@1  & r@5  & r@10  & r@1\%  \\ \hline \hline
		\multirow{4}{*}{CVUSA} & $360^{\circ}$ & 78.11 & 89.46 & 92.90 & 98.50    & 48.38 & 66.73 & 73.82 & 90.93   & 16.36 & 30.67 & 37.78 & 63.82    & 9.26  & 19.82 & 26.08 & 52.45     \\
		& $180^{\circ}$ & 74.30 & 87.37 & 91.25 & 98.27    & 48.53 & 68.47 & 75.63 & 93.02   & 17.91 & 33.51 & 41.24 & 67.40    & 10.42 & 22.43 & 29.06 & 55.98  \\
		& $90^{\circ}$  & 60.13 & 76.23 & 82.40 & 96.21    & 38.98 & 57.94 & 66.38 & 89.39   & 16.19 & 31.44 & 39.85 & 71.13    & 10.15 & 22.02 & 29.50 & 62.78  \\
		& $70^{\circ}$  & 51.65 & 69.56 & 76.09 & 93.75    & 31.79 & 49.99 & 58.45 & 85.69   & 12.74 & 26.78 & 34.73 & 68.00    & 8.78 & 19.90 & 27.30 & 61.20  \\ \hline
		\multirow{4}{*}{CVACT\_val} & $360^{\circ}$ & 72.91 & 85.70 & 88.88 & 95.28    & 44.43 & 63.23 & 69.73 & 87.09   & 13.26 & 26.62 & 33.14 & 60.33    & 6.70  & 16.18 & 21.77 & 48.05     \\
		& $180^{\circ}$ & 72.87 & 85.68 & 88.97 & 95.67    & 49.12 & 67.83 & 74.18 & 89.93   & 17.13 & 33.68 & 41.55 & 67.99    & 10.00 & 22.10 & 28.97 & 56.26  \\
		& $90^{\circ}$  & 64.00 & 78.11 & 82.77 & 94.10    & 43.42 & 61.17 & 68.22 & 86.80   & 18.11 & 33.34 & 40.94 & 68.05    & 11.14 & 23.65 & 31.34 & 59.73  \\
		& $70^{\circ}$  & 55.73 & 71.63 & 77.35 & 92.02    & 34.92 & 52.62 & 60.52 & 83.31   & 13.86 & 27.81 & 35.24 & 64.14    & 9.29 & 20.72 & 27.13 & 57.08  \\\bottomrule
		
	\end{tabular}
\end{table*}

\begin{figure}[!tb]
	\centering
	\includegraphics[width=0.14\linewidth]{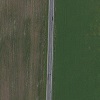}
	\includegraphics[width=0.14\linewidth]{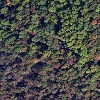}
	\includegraphics[width=0.14\linewidth]{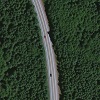}
	\caption{Examples of symmetric scenes (aerial images). At these locations, it is hard to determine the orientation (azimuth angle) of a ground image if it only contains a small sector of the aerial image.}
	\label{fig:symmetric_scene}
\end{figure}

\begin{figure*}[!ht]
	\centering
	\subfigure[FoV=$360^{\circ}$]{
		\begin{minipage}[b]{\linewidth}
			\begin{minipage}[b]{0.07\linewidth}
				\centering
				\includegraphics[width=\linewidth]{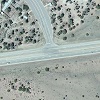}\\
				\includegraphics[width=\linewidth]{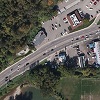}\\
				% \subcaption{Aerial}
				\vspace{-1mm}
				\centerline{\footnotesize Aerial}
			\end{minipage}
			\hspace{-1mm}
			\begin{minipage}[b]{0.21\linewidth}
				\centering
				\includegraphics[width=\linewidth,height=0.3333\linewidth]{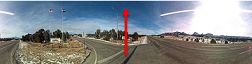}\\
				\includegraphics[width=\linewidth,height=0.3333\linewidth]{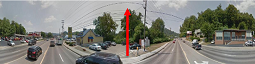}\\
				\vspace{-1mm}
				\centerline{\footnotesize Ground}
			\end{minipage}
			\hspace{-1mm}
			\begin{minipage}[b]{0.21\linewidth}
				\centering
				\scalebox{1.0}[1.0]{\includegraphics[trim={0mm 0mm 0mm 3mm}, clip, width=\linewidth,height=0.3333\linewidth]{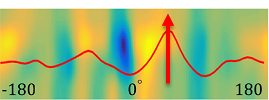}}\\
				\scalebox{1.0}[1.0]{\includegraphics[trim={0mm 0mm 0mm 3mm}, clip, width=\linewidth,height=0.3333\linewidth]{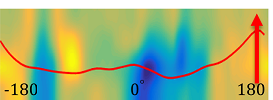}}\\
				\vspace{-1mm}
				\centerline{\footnotesize Similarity Curve}
			\end{minipage}
			\begin{minipage}[b]{0.07\linewidth}
				\centering
				\includegraphics[width=\textwidth]{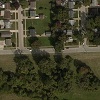}\\
				\includegraphics[width=\textwidth]{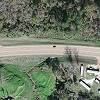}\\
				\vspace{-1mm}
				\centerline{\footnotesize Aerial}
			\end{minipage}
			\hspace{-1mm}
			\begin{minipage}[b]{0.21\linewidth}
				\centering
				\includegraphics[width=\linewidth,height=0.3333\linewidth]{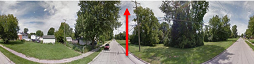}\\
				\includegraphics[width=\linewidth,height=0.3333\linewidth]{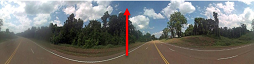}\\
				\vspace{-1mm}
				\centerline{\footnotesize Ground}
			\end{minipage}
			\hspace{-1mm}
			\begin{minipage}[b]{0.21\linewidth}
				\centering
				\scalebox{1.0}[1.0]{\includegraphics[trim={0mm 0mm 0mm 3mm}, clip, width=\linewidth,height=0.3333\linewidth]{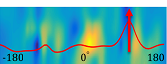}}\\
				\scalebox{1.0}[1.0]{\includegraphics[trim={0mm 0mm 0mm 3mm}, clip, width=\linewidth,height=0.3333\linewidth]{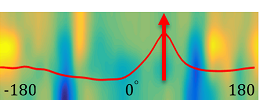}}\\
				\vspace{-1mm}
				\centerline{\footnotesize Similarity Curve}
			\end{minipage}\\
			\vspace{-1em}
		\end{minipage}
		\label{fig: visual_orien_360}
	}\\
	\subfigure[FoV=$180^{\circ}$]{
		\begin{minipage}[b]{\linewidth}
			\begin{minipage}[b]{0.07\linewidth}
				\centering
				\includegraphics[width=\linewidth]{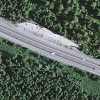}\\
				\includegraphics[width=\linewidth]{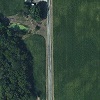}\\
				\vspace{-1mm}
				\centerline{\footnotesize Aerial}
			\end{minipage}
			\hspace{-1mm}
			\begin{minipage}[b]{0.21\linewidth}
				\centering
				\includegraphics[height=0.3333\linewidth]{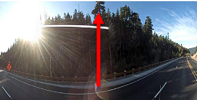}\\
				\includegraphics[height=0.3333\linewidth]{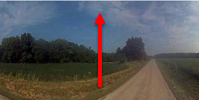}\\
				\vspace{-1mm}
				\centerline{\footnotesize Ground}
			\end{minipage}
			\hspace{-1mm}
			\begin{minipage}[b]{0.21\linewidth}
				\centering
				\scalebox{1.0}[1.0]{\includegraphics[trim={0mm 0mm 0mm 3mm}, clip, width=\linewidth,height=0.3333\linewidth]{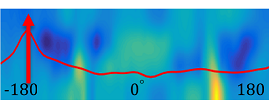}}\\
				\scalebox{1.0}[1.0]{\includegraphics[trim={0mm 0mm 0mm 3mm}, clip, width=\linewidth,height=0.3333\linewidth]{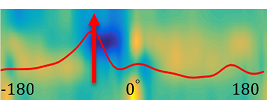}}\\
				\vspace{-1mm}
				\centerline{\footnotesize Similarity Curve}
			\end{minipage}
			\begin{minipage}[b]{0.07\linewidth}
				\centering
				\includegraphics[width=\textwidth]{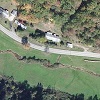}\\
				\includegraphics[width=\textwidth]{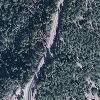}\\
				\vspace{-1mm}
				\centerline{\footnotesize Aerial}
			\end{minipage}
			\hspace{-1mm}
			\begin{minipage}[b]{0.21\linewidth}
				\centering
				\includegraphics[height=0.3333\linewidth]{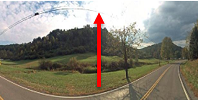}\\
				\includegraphics[height=0.3333\linewidth]{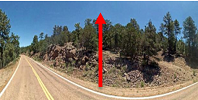}\\
				\vspace{-1mm}
				\centerline{\footnotesize Ground}
			\end{minipage}
			\hspace{-1mm}
			\begin{minipage}[b]{0.21\linewidth}
				\centering
				\scalebox{1.0}[1.0]{\includegraphics[trim={0mm 0mm 0mm 3mm}, clip, width=\linewidth,height=0.3333\linewidth]{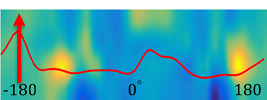}}\\
				\scalebox{1.0}[1.0]{\includegraphics[trim={0mm 0mm 0mm 3mm}, clip, width=\linewidth,height=0.3333\linewidth]{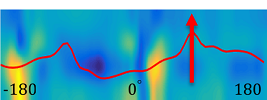}}\\
				\vspace{-1mm}
				\centerline{\footnotesize Similarity Curve}
			\end{minipage}\\
			\vspace{-1em}
		\end{minipage}
		\label{fig: visual_orien_180}}\\
	\subfigure[FoV=$90^{\circ}$]{
		\begin{minipage}[b]{\linewidth}
			\begin{minipage}[b]{0.07\linewidth}
				\centering
				\includegraphics[width=\linewidth]{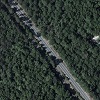}\\
				\includegraphics[width=\linewidth]{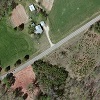}\\
				% \subcaption{Aerial}
				\vspace{-1mm}
				\centerline{\footnotesize Aerial}
			\end{minipage}
			\hspace{-1mm}
			\begin{minipage}[b]{0.21\linewidth}
				\centering
				\includegraphics[height=0.3333\linewidth]{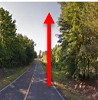}\\
				\includegraphics[height=0.3333\linewidth]{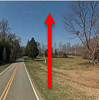}\\
				\vspace{-1mm}
				\centerline{\footnotesize Ground}
			\end{minipage}
			\hspace{-1mm}
			\begin{minipage}[b]{0.21\linewidth}
				\centering
				\scalebox{1.0}[1.0]{\includegraphics[trim={0mm 0mm 0mm 3mm}, clip, width=\linewidth,height=0.3333\linewidth]{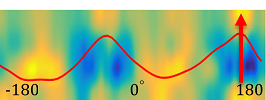}}\\
				\scalebox{1.0}[1.0]{\includegraphics[trim={0mm 0mm 0mm 3mm}, clip, width=\linewidth,height=0.3333\linewidth]{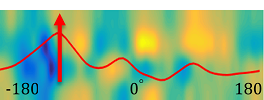}}\\
				\vspace{-1mm}
				\centerline{\footnotesize Similarity Curve}
			\end{minipage}
			\begin{minipage}[b]{0.07\linewidth}
				\centering
				\includegraphics[width=\textwidth]{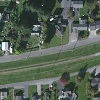}\\
				\includegraphics[width=\textwidth]{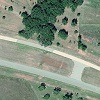}\\
				\vspace{-1mm}
				\centerline{\footnotesize Aerial}
			\end{minipage}
			\hspace{-1mm}
			\begin{minipage}[b]{0.21\linewidth}
				\centering
				\includegraphics[height=0.3333\linewidth]{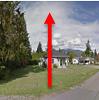}\\
				\includegraphics[height=0.3333\linewidth]{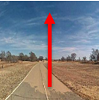}\\
				\vspace{-1mm}
				\centerline{\footnotesize Ground}
			\end{minipage}
			\hspace{-1mm}
			\begin{minipage}[b]{0.21\linewidth}
				\centering
				\scalebox{1.0}[1.0]{\includegraphics[trim={0mm 0mm 0mm 3mm}, clip, width=\linewidth,height=0.3333\linewidth]{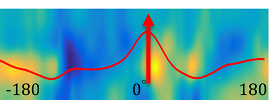}}\\
				\scalebox{1.0}[1.0]{\includegraphics[trim={0mm 0mm 0mm 3mm}, clip, width=\linewidth,height=0.3333\linewidth]{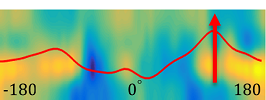}}\\
				\vspace{-1mm}
				\centerline{\footnotesize Similarity Curve}
			\end{minipage}\\
			\vspace{-1em}
		\end{minipage}
		\label{fig: visual_orien_90}
	}\\
	\subfigure[FoV=$70^{\circ}$]{
		\begin{minipage}[b]{\linewidth}
			\begin{minipage}[b]{0.07\linewidth}
				\centering
				\includegraphics[width=\linewidth]{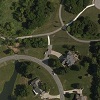}\\
				\includegraphics[width=\linewidth]{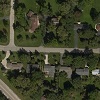}\\
				\vspace{-1mm}
				\centerline{\footnotesize Aerial}
			\end{minipage}
			\hspace{-1mm}
			\begin{minipage}[b]{0.21\linewidth}
				\centering
				\includegraphics[height=0.3333\linewidth]{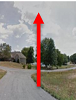}\\
				\includegraphics[height=0.3333\linewidth]{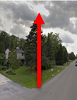}\\
				\vspace{-1mm}
				\centerline{\footnotesize Ground}
			\end{minipage}
			\hspace{-1mm}
			\begin{minipage}[b]{0.21\linewidth}
				\centering
				\scalebox{1.0}[1.0]{\includegraphics[trim={0mm 0mm 0mm 3mm}, clip, width=\linewidth,height=0.3333\linewidth]{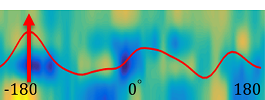}}\\
				\scalebox{1.0}[1.0]{\includegraphics[trim={0mm 0mm 0mm 3mm}, clip, width=\linewidth,height=0.3333\linewidth]{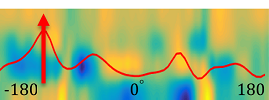}}\\
				\vspace{-1mm}
				\centerline{\footnotesize Similarity Curve}
			\end{minipage}
			\begin{minipage}[b]{0.07\linewidth}
				\centering
				\includegraphics[width=\textwidth]{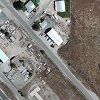}\\
				\includegraphics[width=\textwidth]{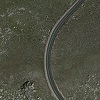}\\
				\vspace{-1mm}
				\centerline{\footnotesize Aerial}
			\end{minipage}
			\hspace{-1mm}
			\begin{minipage}[b]{0.21\linewidth}
				\centering
				\includegraphics[height=0.3333\linewidth]{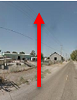}\\
				\includegraphics[height=0.3333\linewidth]{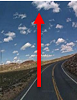}\\
				\vspace{-1mm}
				\centerline{\footnotesize Ground}
			\end{minipage}
			\hspace{-1mm}
			\begin{minipage}[b]{0.21\linewidth}
				\centering
				\scalebox{1.0}[1.0]{\includegraphics[trim={0mm 0mm 0mm 3mm}, clip, width=\linewidth,height=0.3333\linewidth]{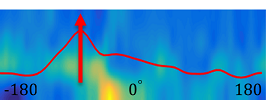}}\\
				\scalebox{1.0}[1.0]{\includegraphics[trim={0mm 0mm 0mm 3mm}, clip, width=\linewidth,height=0.3333\linewidth]{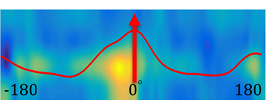}}\\
				\vspace{-1mm}
				\centerline{\footnotesize Similarity Curve}
			\end{minipage}\\
			\vspace{-1em}
		\end{minipage}
		\label{fig: visual_orien_70}}\\
	\vspace{-0.2em}
	\caption{Visualization of estimated orientations for ground images with varying FoVs. For each ground and aerial pair, we visualize their similarity scores at different azimuth angles as a red curve on the polar-transformed aerial features. As indicated by the arrows on the similarity curves, the positions of similarity maxima in the curves correspond to the orientation of ground images.
		% The correlation results between ground and aerial images at different azimuth angles are visualized by red curves on polar-transformed aerial features. As indicated by the arrows on the similarity curves, the positions of similarity maxima in the curves corresponds to the orientation of ground images.
		% In each of the subfigures, ``Aerial'' means aerial images and ``Ground'' means ground images. The similarity score (correlation results) between them at different azimuth angles are visualized by ``red curves'' on the polar-transformed aerial features. As indicated by the arrows on the similarity curves, the positions of similarity maxima in the curves corresponds to the orientation of ground images.
		% The polar-transformed aerial features and left are aerial images and middle are ground images. We visualize the polar transformed-aerial features and the correlation results (red curves) between ground and aerial features in the right column. The positions of maximum values in the curves corresponds to the orientations of ground images. 
	}
	\label{fig:visual_orien}
\end{figure*}

We provide additional visualization of estimating orientations for ground images with $360^{\circ}$, $180^{\circ}$, $90^{\circ}$ and $70^{\circ}$ FoV in Figure \ref{fig:visual_orien}. As illustrated in Figure \ref{fig:visual_orien}, our Dynamic Similarity Matching (DSM) module is able to estimate the orientation of ground images with varying FoVs.

When a camera has a small FoV, it suffers high ambiguity to determine the orientation by matching the ground image to its corresponding aerial one. 
We illustrate an example in Figure \ref{fig: visual_orien_90} where the error of orientation estimation for a road can be $180^{\circ}$. 
As seen in the first instance of Figure \ref{fig: visual_orien_90}, 
the road occupies a large portion of the ground image. While in the aerial image, the road is symmetric in respect to the image center (\ie, camera location). Thus there are two peaks in the similarity curve.

If the peak on the left is taller than the peak on the right, the estimated orientation will be wrong.
Figure \ref{fig:symmetric_scene} provides another three examples where scene contents are similar in multiple directions. At these locations, it is hard to determine the orientation of a ground camera which has a small FoV while the estimated location is correct.

% Figure \ref{fig: CVUSA_unkown_orien_FOV_localization_example} presents some examples of joint location and orientation estimation for ground images with unknown orientation and different FoV.

\section{Time Efficiency}
In order to improve the time efficiency, we compute the correlation in our DSM module by using Fast Fourier Transform during the inference process. To be specific, we store the Fourier coefficients of aerial features in the database, and calculate the Fourier coefficients of the ground feature in the forward pass. 
By doing so, the computation flops of the correlation are $13NHWC$ (including $4NHWC$ flops for coefficients multiplication in the spectral domain, and $1.5NHCW\log_2 W$ flops for the inverse Fast Fourier Transform), where $H$, $W$ and $C$ is the height, width and channel number of the global feature descriptor of an aerial image, $N$ is the number of database aerial images, and $W=64$ in our method.  
In contrast to conducting correlation in the spatial domain where the computation flops are $2NHW^2C$, the computation time is reduced by a factor of $\frac{1}{10}$ ($\frac{13NHWC}{2NHW^2C}\approx \frac{1}{10}$).

We conduct the retrieval process of a query image on a 3.70 GHz i7 CPU system, and the codes are implemented in Python3.6.
For a ground panorama with unknown orientation, it takes an average time of $0.15s$ for retrieving its aerial counterpart from a database containing 8884 reference images. This demonstrates the efficiency of the proposed algorithm.

\section{Trainable Parameters}

Since the authors of CVM-NET \cite{Hu_2018_CVPR}, Liu \& Li \cite{Liu_2019_CVPR} and CVFT \cite{shi2019optimal} provide the source code of their works, we compare the trainable parameters and model size of our network with their methods in Table \ref{tab: compare_parameters}. Our network not only outperforms the state-of-the-art but also is more compact, facilitating the deployment of our network.

\begin{table}[!ht]
	\setlength{\tabcolsep}{10pt}
	\centering
	\caption{\small Comparison of trainable parameters and model size with recent methods. }
	\begin{tabular}{c | c c}
		\toprule
		Methods & \# Parameters & Model size  \\ \hline \hline
		
		CVM-NET  \cite{Hu_2018_CVPR}          & 160,311,424 & 1.8G     \\  
		Liu \& Li \cite{Liu_2019_CVPR}        & 30,695,808 & 369.6M      \\ 
		CVFT \cite{shi2019optimal}            & 26,814,657 & 336.8M  \\ 
		Ours                            & 14,472,864 & 244.8M   \\ 
		\bottomrule
		% \vspace{-1em}
	\end{tabular}
	\label{tab: compare_parameters}
\end{table}

\end{document}